\pdfoutput=1
\documentclass[10pt,twocolumn,letterpaper]{article}

\usepackage[pagenumbers]{cvpr} %

\makeatletter
\renewcommand{\cvprsection}{\@startsection{section}{1}{\z@}%
  {-8pt plus -2pt minus -2pt}{5pt}{\large\bf}}
\renewcommand{\cvprsubsection}{\@startsection{subsection}{2}{\z@}%
  {-6pt plus -2pt minus -2pt}{3pt}{\elvbf}}
\makeatother

\usepackage{tikz}
\usetikzlibrary{arrows.meta}

\usepackage{multirow,pifont}

\definecolor{GrapeBrand}{HTML}{8E75C6}
\definecolor{benchimage}{HTML}{8C8C8C}
\definecolor{benchvideo}{HTML}{F06450}
\definecolor{benchdriving}{HTML}{159B73}
\colorlet{benchours}{GrapeBrand}
\DeclareRobustCommand{\benchdot}[2][0.9em]{\makebox[#1][l]{%
  \begingroup
  \setbox0=\hbox{X}%
  \raisebox{0.5\ht0}[0pt][0pt]{%
    \tikz[baseline=0pt]{\fill[#2] (0,0) circle (1.6pt);}}%
  \endgroup}}

\AtBeginDocument{%
  \pdfstringdefDisableCommands{}%
}

\definecolor{GrapeFinding}{HTML}{7460C2}

\usepackage{xspace}
\newcommand{\method}{\textsc{Morse}\xspace}
\newcommand{\editor}{Editor\xspace}
\newcommand{\solver}{Solver\xspace}

\definecolor{morsestep1}{HTML}{E5675C}
\definecolor{morsestep2}{HTML}{DB9A1F}
\definecolor{morsestep3}{HTML}{4FA765}
\definecolor{morsestep4}{HTML}{4285D6}
\DeclareRobustCommand{\morsestep}[1]{%
  \begingroup\setbox0=\hbox{X}%
  \raisebox{0.5\ht0}[0pt][0pt]{%
    \tikz[baseline=(step.center)]{%
      \node[circle,fill=morsestep#1,text=white,inner sep=0pt,outer sep=0pt,
        minimum size=9.4pt,font=\bfseries\fontsize{6.7}{6.7}\selectfont]
        (step) {#1};}}%
  \endgroup\hspace{0.3em}}

\definecolor{morseproprietary}{HTML}{3B6FB6} %
\definecolor{morsepretrained}{HTML}{C3A027} %
\definecolor{morsestreaming}{HTML}{B84D85} %
\definecolor{morsersi}{HTML}{8056AA}

\definecolor{morsereference}{HTML}{606060}
\newcommand{\morseref}[1]{{\hypersetup{citecolor=morsereference}\color{morsereference}#1}}

\DeclareRobustCommand{\modelicon}[1]{%
  \begingroup\setbox0=\hbox{X}%
  \raisebox{\dimexpr(\ht0-\height)/2\relax}[0pt][0pt]{%
    \includegraphics[height=0.92em,width=0.95em,keepaspectratio]{logos/#1}}%
  \endgroup\hspace{0.35em}}
\DeclareRobustCommand{\rsilabel}{\textcolor{morsersi}{\textsf{RSI}}}

\newcommand{\morsegroupseparator}{\rlap{\raisebox{0.93em}[0pt][0pt]{%
  \tikz[overlay,baseline=0pt]{\draw[black!35,line width=0.22pt,
    dash pattern=on 1.5pt off 1.5pt] (0,0) -- (\textwidth,0);}}}}

\makeatletter
\definecolor{morseoursband}{HTML}{EDE5FB}
\newcommand{\morsehighlight}{\rlap{\color{morseoursband}%
  \vrule width\textwidth height\ht\@arstrutbox depth\dp\@arstrutbox\relax}}
\newcommand{\morsehighlightcol}{\rlap{\color{morseoursband}%
  \vrule width\columnwidth height\ht\@arstrutbox depth\dp\@arstrutbox\relax}}
\makeatother

\definecolor{morseyes}{HTML}{2E8B57}
\definecolor{morseno}{HTML}{C8473F}
\newcommand{\yesmark}{\textcolor{morseyes}{\ding{51}}}
\newcommand{\nomark}{\textcolor{morseno}{\ding{55}}}
\newlength{\morsenumw}
\newcommand{\tdelta}[2]{\makebox[2.1em][l]{\hspace{0.12em}\raisebox{-0.22em}[0pt][0pt]{\fontsize{6}{6}\selectfont\color{#1}#2}}}
\newcommand{\tv}[1]{\makebox[1.8em][r]{#1}\makebox[2.1em][l]{}}
\newcommand{\tup}[2]{\makebox[1.8em][r]{#1}\tdelta{morseyes}{+#2}}

\newcommand{\thd}[1]{\makebox[1.8em][c]{#1}\makebox[2.1em][l]{}}

\definecolor{cvprblue}{rgb}{0.21,0.49,0.74}
\usepackage[pagebackref,breaklinks,colorlinks,allcolors=cvprblue]{hyperref}
\definecolor{grapeurl}{HTML}{EC008C}
\hypersetup{urlcolor=grapeurl}
\let\GrapeOriginalTTDefault\ttdefault
\usepackage[varl,noupquote]{inconsolata}
\let\ttdefault\GrapeOriginalTTDefault

\AtBeginDocument{%
  \crefname{figure}{Figure}{Figures}%
  \Crefname{figure}{Figure}{Figures}%
  \crefname{table}{Table}{Tables}%
  \Crefname{table}{Table}{Tables}%
}

\usepackage{graphicx}

\usepackage{graphicx,booktabs,multirow,array}
\usepackage{xcolor,colortbl}
\usepackage{tikz}

\def\confName{CVPR}
\def\confYear{2027}

\title{\textsc{Morse}: Optimizing Streaming Memory Design via Open-World Auto-Research}

\author{Ce Zhang$^{1*}$ \quad Jianshu Zhang$^{2*}$ \quad Jinxi He$^{1}$ \quad Xiyuan Yang$^{3}$ \\
Pan Wang$^{4}$ \quad Han Liu$^{2}$ \quad Katia Sycara$^{1}$ \quad Yaqi Xie$^{3}$ \\[3pt]
$^{1}$CMU \quad $^{2}$Northwestern \quad $^{3}$UIUC \quad $^{4}$Pitt \quad $^{*}$Equal contribution \\[3pt]
Project page: \url{https://zhangce01.github.io/morse/}
}

\begin{document}
\maketitle
\suppressfloats[t]
\begin{figure}[t]
  \centering
  \includegraphics[width=\linewidth]{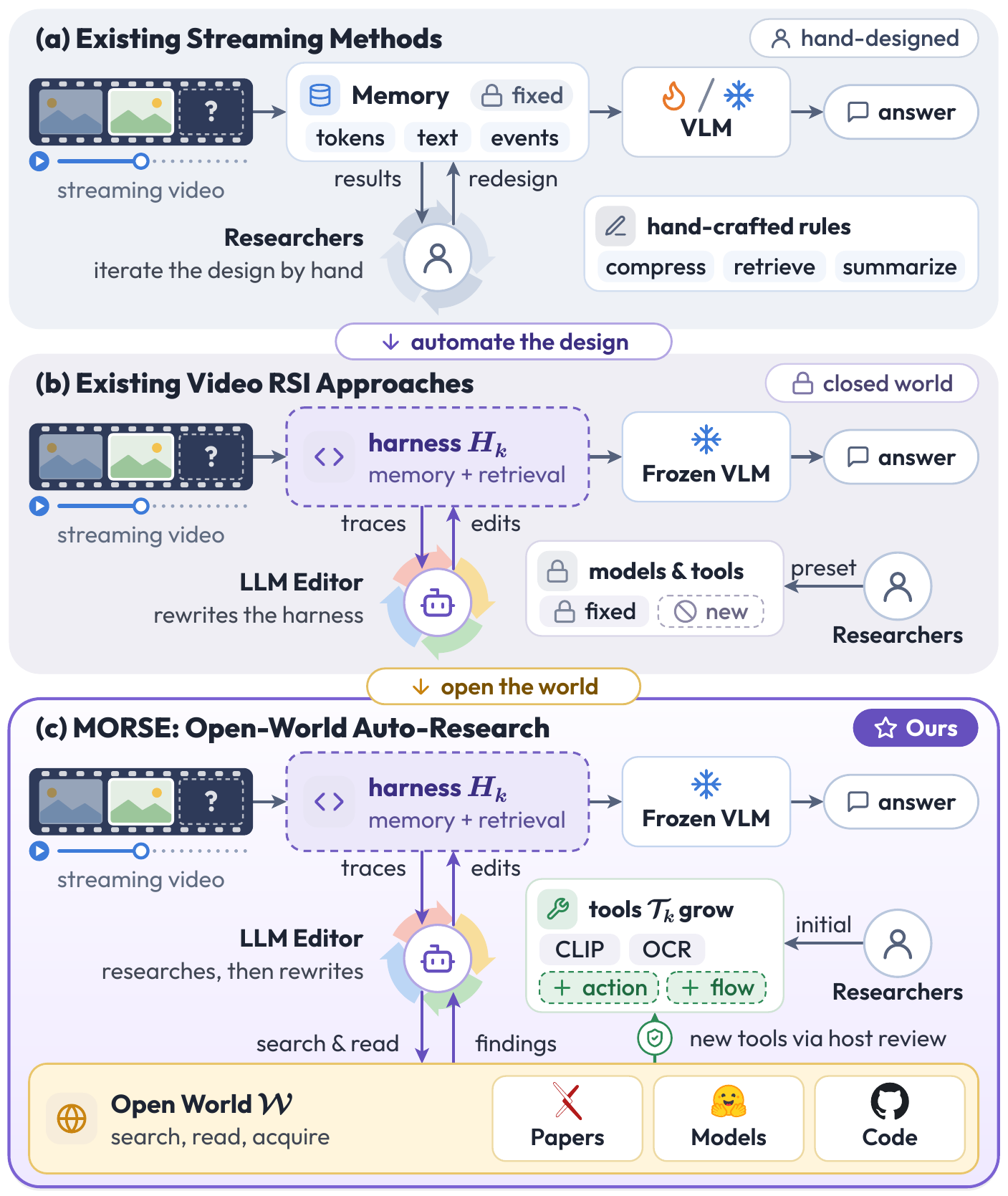}
  \caption{\textbf{From manual design to open-world harness evolution.} \textbf{(a)} Researchers design and revise
  the memory and retrieval pipelines of streaming systems~\cite{yang2025streammem,xie2026fluxmem,di2025rekv,zhang2026streamscout}.
  \textbf{(b)} Video recursive self-improvement (RSI) methods~\cite{luo2026videorsi,lin2026videoharnesssrsi,liu2026memevo}
  automate harness revision from execution feedback within a preconfigured set of models and tools.
  \textbf{(c)} \method{} makes online research part of this loop. An LLM-based \editor{} investigates observed failures
  through papers, model cards, and code, then adapts existing methods and requests missing perception tools.
  Host-reviewed tool acquisition lets the harness and its available capabilities evolve together around a frozen
  answering VLM.}
  \label{fig:teaser}
\end{figure}
\begin{abstract}
A streaming video system must decide in a single pass what evidence to retain under bounded frame storage, before it knows which details a future question will require. Recent recursive self-improvement methods automate the design of the harness that stores and retrieves this evidence. They revise the harness in response to failures, but the editor must still devise each repair from its prior knowledge and a predefined set of tools. A failure that recurs across videos, however, may point to a problem that prior work has already addressed. We introduce \method{} (Memory Optimization with Recursive Self-Evolvement), an open-world auto-research framework that uses recurring failures to guide an online search for solutions. Its editor groups failures in training traces by cause, then searches papers, model cards, and code for methods it can adapt or pretrained perception models it can use. When the editor introduces a new perceptual capability, it can also revise how the harness stores and retrieves the resulting evidence for a frozen video-language model. On held-out  validation streams, \method{} first selects revisions that improve answer quality, then seeks faster revisions that preserve those gains within a specified tolerance. Across three streaming video benchmarks, \method{} consistently achieves the highest answer accuracy within each frozen-backbone comparison. On OVO-Bench, our Qwen3-VL-8B harness answers queries in 0.30 seconds on average while achieving 75.1\% accuracy, 3.1 percentage points ahead of the next-best evaluated harness using the same frozen backbone.
\end{abstract}

\section{Introduction}
\label{sec:intro}

Streaming video understanding requires answering questions about an ongoing video stream using only observations available at query time~\cite{niu2025ovobench,lin2026streamingbench}. In this single-pass setting, the system has limited space to store frames and must decide what to preserve before it knows which details a future question will require. We use the term streaming \emph{harness} for the policies that maintain this memory and retrieve evidence for a video-language model (VLM)~\cite{lin2026videoharnesssrsi}. As frames arrive, its write policy selects which observations to retain and how to represent and organize them in memory. When a question arrives, its read policy uses the question to retrieve relevant evidence and assemble the context for the VLM. These policies must be designed together. The write policy determines which details remain available, and even a retained detail can help answer a question only if the read policy can locate it within the latency budget.

As illustrated in \mbox{\cref{fig:teaser}(a)}, existing streaming systems use researcher-designed policies for memory compression, temporal organization, and query-aware retrieval~\cite{yang2025streammem,xie2026fluxmem,di2025rekv,zhang2026streamscout}. Recent recursive self-improvement (RSI) methods automate parts of this design process by analyzing failures and revising the harness code~\cite{luo2026videorsi,lin2026videoharnesssrsi,liu2026memevo}. In the loop shown in \mbox{\cref{fig:teaser}(b)}, the editor draws on its prior knowledge to devise revisions using a predefined set of models and tools. However, this limits the search to solutions the editor can recall or develop on its own. Faced with a failure that recurs across videos, a human researcher would consult the literature to see whether others have already addressed the underlying problem. Even work on a different task may offer a method to adapt or a pretrained model that provides a capability the harness lacks. An editor could follow the same approach, using the failures it diagnoses to guide an online search for solutions it can adapt to the harness.

Motivated by this, we introduce \method{} (Memory Optimization with Recursive Self-Evolvement), an \emph{open-world auto-research} framework that lets an editor read papers online and adapt relevant methods to its own streaming video harness.\footnote{To prevent evaluation data leakage, we block online access to benchmark data and answers and exclude tools trained or fine-tuned on evaluation videos or annotations.} Specifically, \method{} has two roles. The streaming \solver{} executes the harness, updating memory as frames arrive and retrieving evidence for a frozen VLM when a question is asked. The LLM-based \editor{} inspects the \solver{}'s training traces, identifies recurring failures, and searches papers, model pages, and code for approaches that address their causes. It translates its findings into changes to the memory and retrieval policies. When a failure calls for a new perceptual capability, it can request a pretrained model and adapt these policies to use its outputs after review and integration. As shown in \mbox{\cref{fig:teaser}(c)}, online research informs both program revisions and the expansion of available perception tools, allowing the policies and capabilities to evolve together. Held-out validation guides two stages of evolution: first improving answer quality, then reducing answer latency while bounding quality loss. Both stages enforce answer-latency and write-load limits, and the selected harness becomes the starting point for the next round. %

Starting from a simple harness, \method{} improves held-out validation accuracy from 68.3\% to 79.4\% through recursive evolution.
On held-out test benchmarks with Qwen3-VL-8B, the final harness achieves 75.1\% on OVO-Bench (RT+BT), 78.9\% on
StreamingBench, and 71.6\% on StreamBench, exceeding the next-best evaluated harness using the same frozen backbone by
3.1, 2.7, and 6.3 percentage points, respectively. These gains span backward tracing, contextual reasoning,
and memory search. Beyond accuracy, \method{} answers 6.0$\times$ and 3.5$\times$ faster than MemEvo and OASIS.
Its harness transfers unchanged to LLaVA-OneVision-2-8B without further evolution. \looseness=-1

\begingroup
\setlength{\parskip}{0pt}
Our contributions are summarized as follows:
\begin{itemize}[topsep=3pt,itemsep=3pt,partopsep=0pt]
    \item We introduce \method{}, an open-world auto-research framework that uses streaming memory failures to guide online research, enabling recursive improvements beyond predefined designs and capabilities.

    \item We expand the search space of harness evolution from program code to perceptual capabilities, so that memory and retrieval policies evolve together with the perception tools they use while the answering VLM stays frozen.

    \item Experiments on OVO-Bench, StreamingBench, and StreamBench show that \method{} outperforms existing streaming methods and harness RSI baselines across answering models, and ablations quantify the contribution of external research and of the expanded capability space.
\end{itemize}
\endgroup

\begin{figure*}[t]
  \centering
  \includegraphics[width=\linewidth]{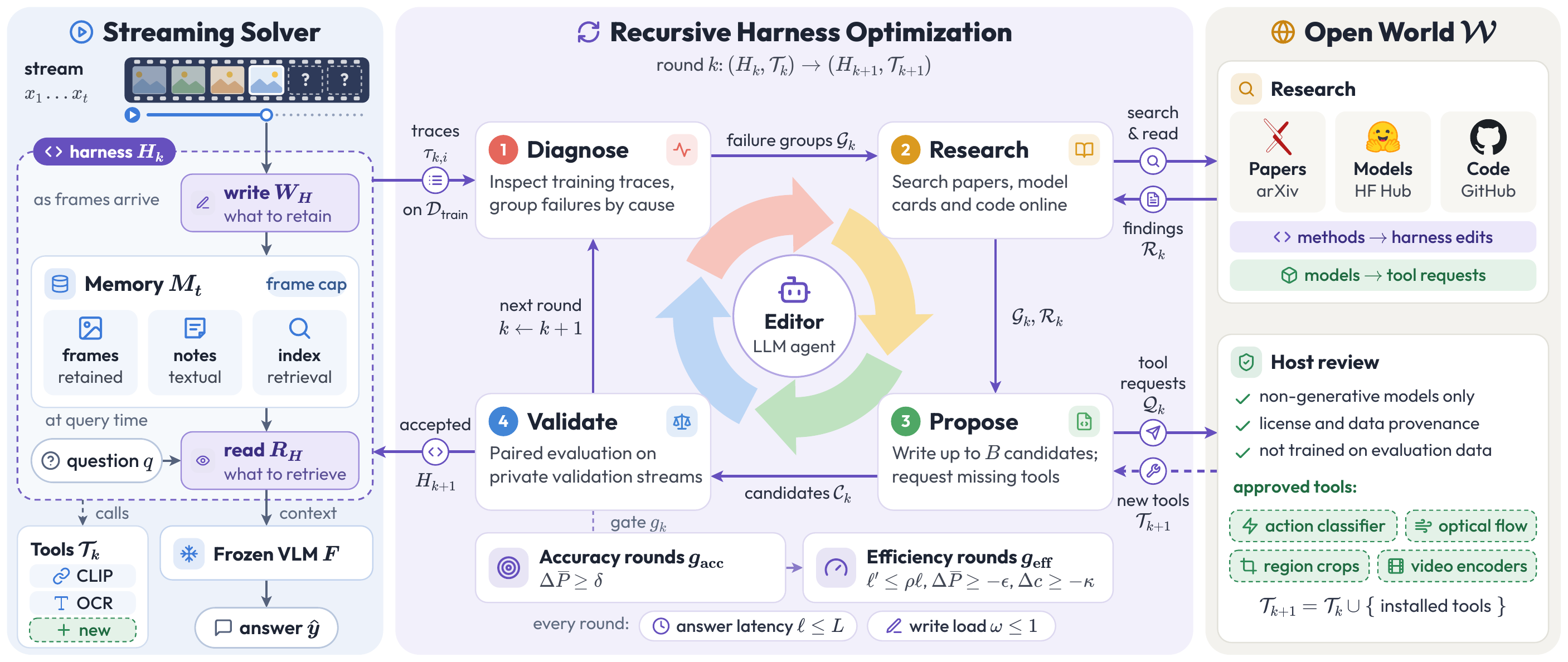}
  \caption{\textbf{The \method{} optimization loop.} \emph{Left:} the harness $H_k$ writes observations to memory
  through $W_H$ and retrieves evidence through $R_H$ for the frozen VLM $F$.  \emph{Middle:} the LLM-based \editor{} diagnoses errors and
  costly operations from training traces, researches solutions, proposes executable revisions, and validates them on
  private streams. \emph{Right:} papers, model cards, and code inform program edits and tool requests; approved
  tools become callable after host integration. \emph{Below:} accuracy rounds improve answer quality; efficiency
  rounds reduce latency with bounded quality loss relative to both the parent and the stage-entry harness.}
  \label{fig:overview}
\end{figure*}
\section{Related Work}
\label{sec:related}

\noindent\textbf{Streaming Video Understanding.}
Streaming video understanding requires systems to preserve useful history as frames arrive and retrieve evidence
when a question is posed, without access to future observations. Existing methods address these demands through
two complementary directions. Training-based systems learn continuous interaction and memory mechanisms within
the model~\cite{chen2024videollmonline,zhang2025flashvstream,zeng2025streamforest}. Inference-time approaches build
on existing backbones by designing how evidence is stored and accessed: StreamMem and FluxMem compress or organize
visual history~\cite{yang2025streammem,xie2026fluxmem}, while ReKV and StreamScout retrieve cached representations
or acquire query-relevant evidence~\cite{di2025rekv,zhang2026streamscout}. These methods offer different trade-offs
between retaining history and answering efficiently, but their overall memory structures and processing pipelines
are specified by researchers, even when individual decisions are adaptive or learned. 
\method{} automates the design of these policies, using observed failures to revise an executable memory and
retrieval program around a frozen answering model.

\vspace{3pt} \noindent\textbf{Harness Recursive Self-Improvement.}
Recursive self-improvement turns evaluation feedback into program changes that persist beyond an individual task.
LLM-guided evolution applies this principle to executable programs, agent workflows, and memory architectures
through repeated proposal, evaluation, and selection~\cite{hu2025adas,zhang2025aflow,novikov2025alphaevolve,zhang2026memevolve}.
For video understanding, recent methods evolve context-construction and tool-use programs around frozen perception
models~\cite{xu2026videoharnessrsi,ye2026videoresearcher}, with failure diagnosis and controlled feedback guiding
the search for more generalizable harness revisions~\cite{luo2026videorsi,lin2026videoharnesssrsi}.
Closest to our streaming setting, MemEvo searches causal memory programs with extensible operators within a fixed
model environment~\cite{liu2026memevo}.
DeepEvolve incorporates online research into program evolution, using external knowledge to guide algorithm
revisions beyond what the editor can propose from prior knowledge alone~\cite{liu2025deepevolve}.
\method{} combines online research with host-reviewed acquisition of perception models, evolving both streaming
harness code and its available capabilities under answer-latency and write-cost constraints.

\begingroup
\setlength{\abovedisplayskip}{6pt plus 2pt minus 2pt}
\setlength{\belowdisplayskip}{6pt plus 2pt minus 2pt}
\setlength{\abovedisplayshortskip}{0pt plus 2pt}
\setlength{\belowdisplayshortskip}{4pt plus 2pt minus 2pt}
\section{Method}
\label{sec:method}

\method{} uses these failures to guide program revision and external research. As shown in \cref{fig:overview},
a \solver{} answers with a frozen VLM, while an LLM-based \editor{} develops its memory and retrieval policies,
first improving answer quality and then reducing execution cost.

\subsection{Problem Formulation}
\label{sec:setup}

A video stream supplies frames $x_1,x_2,\ldots$ sequentially. A question $q$ arriving at time $t$ must be answered
from $x_{1:t}$; future questions are unknown when memory is written. In our single-pass setting, memory $M_t$
comprises a recent-frame window, a persistent frame store $V_t$ with $|V_t|\le G$, and derived captions, embeddings,
and retrieval indices. Once a frame leaves the window, its pixels are accessible only if retained in $V_t$.
Derived text and vectors have no host-imposed global storage bound and can grow with stream duration.

The harness $H=(W_H,R_H)$ is an executable program whose write policy $W_H$ updates memory as frames arrive
and whose read policy $R_H$ constructs a query-specific context for the answering model:
\begin{equation}
  \begin{aligned}
    M_t &= W_H\big(M_{t-1},x_t;\,\mathcal{T}\big), \\
    \hat{y} &= F\big(q,\;R_H(M_t,q;\,\mathcal{T})\big).
  \end{aligned}
  \label{eq:harness}
\end{equation}
Here, $F$ is the frozen answering VLM and $\mathcal{T}$ contains the callable model services. The harness may change
memory organization, frame selection, retrieval, prompts, and service use while keeping model weights fixed.

During evolution, we estimate call durations with a fixed linear cost model to avoid server-load fluctuations
affecting selection. Let $\ell_i$ denote estimated answer latency for question $i$, including retrieval and VLM
calls, and $w_s$ the estimated write-path seconds on stream $s$ of duration $S_s$. Over $N$ validation questions,
the mean latency and maximum write load are
\begin{equation}
  \ell(H)=\frac{1}{N}\sum_{i=1}^{N}\ell_i,
  \qquad
  \omega(H)=\max_s\frac{w_s}{S_s}.
  \label{eq:cost}
\end{equation}
An admissible harness satisfies $\ell(H)\le L$ and $\omega(H)\le1$, where $L$ is the answer-latency budget,
as well as the frame-storage cap. We first improve answer quality within these limits, then reduce latency with
bounded quality loss. During search, mean correct-option probability serves as a surrogate for discrete accuracy.

\subsection{Recursive Harness Optimization}
\label{sec:loop}

Starting from $(H_0,\mathcal{T}_0)$, the \editor{} repeats four steps. Update $k$ produces $H_{k+1}$ and is plotted
as round $k+1$. Each round starts a fresh session with the current program, training traces, past candidate scores
and diffs, reusable observations, and available services, allowing later attempts to build on earlier evidence.

\vspace{3pt} \noindent \textbf{\morsestep{1}Diagnose.}
The \editor{} examines training trace $\tau_{k,i}$, including stored evidence, retrieval choices, service outputs,
and the VLM's context. These traces and revisits to training frames distinguish missing memory from failed
retrieval or perception. The resulting failure groups $\mathcal{G}_k$ identify targets such as preserving brief
interactions or recovering earlier procedure steps. Efficiency rounds also examine correctly answered questions
for expensive or redundant calls.

\vspace{3pt} \noindent \textbf{\morsestep{2}Research.}
\label{sec:openworld}
For each target, the \editor{} searches papers, model pages, and code in the external collection $\mathcal{W}$,
gathering findings $\mathcal{R}_k$ within a request budget $B_{\mathrm{res}}$. A paper may suggest a better temporal
index, while a model page may reveal an action classifier for interactions missed by frame embeddings.
Interfaces and implementations inform integration. A broker blocks access to benchmark data and question-answer
annotations during research.

\vspace{3pt} \noindent \textbf{\morsestep{3}Propose.}
The \editor{} turns these findings into up to $B$ executable candidates $\mathcal{C}_k$ and tool requests
$\mathcal{Q}_k$. Candidates revise which events to record, how to retrieve them, and how to present evidence.
Requests specify missing capabilities, supporting research, and their integration into these policies. The host
controls execution and service access, reviews training-data provenance, excludes models trained or fine-tuned on
evaluation videos or annotations, and admits only non-generative additions.

Approved tools become callable after host integration. Let $\Delta\mathcal{T}_k$ contain services integrated during
round $k$, including earlier pending requests; then
$\mathcal{T}_{k+1}=\mathcal{T}_k\cup\Delta\mathcal{T}_k$. Candidates may use a service as soon as integration
completes. Research can change both code and perception capabilities.

\vspace{3pt} \noindent \textbf{\morsestep{4}Validate.}
The host evaluates candidates on the same private validation streams and returns aggregate quality and cost
measurements to select subsequent harnesses. The \editor{} sees training questions and traces, but no validation
questions or item-level feedback; test benchmarks remain outside development and selection. Identical model
requests reuse cached responses for consistent paired comparisons under the same timing convention.

\subsection{Accuracy and Efficiency Gates}
\label{sec:gate}

\method{} uses $K_{\mathrm{acc}}$ rounds to improve answer quality, then $K_{\mathrm{eff}}$ rounds to reduce cost,
with $K=K_{\mathrm{acc}}+K_{\mathrm{eff}}$. Following prior work~\cite{luo2026videorsi,liu2026memevo}, selection
uses held-out streams.

\vspace{3pt} \noindent \textbf{Answer quality.}
For $N$ validation questions with gold answers $y_i$ and predictions $\hat{y}_i$, let $p_i$ be the correct-option
probability renormalized over valid option letters at the answer position. We compute
\begin{equation}
  \bar{P}(H)=\frac{1}{N}\sum_{i=1}^{N}p_i,
  \qquad c(H)=\sum_{i=1}^{N}[\hat{y}_i=y_i].
  \label{eq:scores}
\end{equation}
Brackets denote a Boolean indicator. Unlike accuracy $c/N$, $\bar{P}$ detects gains before predictions change.
For candidate $H'$ and parent $H$, let
$\Delta\bar{P}=\bar{P}(H')-\bar{P}(H)$, $\Delta c=c(H')-c(H)$, and $\ell'=\ell(H')$.

\vspace{3pt} \noindent \textbf{Accuracy gate.}
The accuracy gate $g_{\mathrm{acc}}$ requires a probability gain of at least $\delta$:
\begin{equation}
  \Delta\bar{P}\ge\delta.
  \label{eq:gate_acc}
\end{equation}
Relative latency and correct counts are unconstrained.

\vspace{3pt} \noindent \textbf{Efficiency gate.}
The efficiency gate $g_{\mathrm{eff}}$ bounds latency by a fraction $\rho$ of the parent's and limits quality loss:
\begin{equation}
  \ell'\le\rho\,\ell(H),\quad \Delta\bar{P}\ge-\epsilon,\quad \Delta c\ge-\kappa.
  \label{eq:gate_eff}
\end{equation}
We bound per-update losses by $\epsilon$ and $\kappa$, and total losses relative to the stage-entry harness
$H_{\mathrm{acc}}=H_{K_{\mathrm{acc}}}$:
\begin{equation}
  \begin{aligned}
    \bar{P}(H')&\ge\bar{P}(H_{\mathrm{acc}})-\epsilon_{\mathrm{tot}}, \\
    c(H')&\ge c(H_{\mathrm{acc}})-\kappa_{\mathrm{tot}}.
  \end{aligned}
  \label{eq:eff_budget}
\end{equation}
\vspace{3pt} \noindent \textbf{Resource limits.}
Both gates require $\ell(H')\le L$, $\omega(H')\le1$, and completion of all validation streams with unchanged
question counts. Let $g_k$ denote the accuracy gate for $k<K_{\mathrm{acc}}$ and the efficiency gate thereafter.

\vspace{3pt} \noindent \textbf{Merging and selection.}
The \editor{} may reapply saved diffs from rejected candidates to the current head, submitting the combined
program as a new candidate that must pass the gate. Separately, the merge stage combines current-round candidates
that pass the gate and share a parent; each merged program is revalidated.

Let $\mathcal{A}_k$ contain this round's passing candidates and merges, and $\mathcal{U}_k$ the archived,
qualified but unselected candidates. Accuracy selection pools these with the current head,
$\mathcal{S}_k=\{H_k\}\cup\mathcal{A}_k\cup\mathcal{U}_k$; efficiency selection uses only
$\mathcal{S}_k=\{H_k\}\cup\mathcal{A}_k$. After deduplication, we select within numerical tie tolerances:
\begin{equation}
  H_{k+1}=\begin{cases}
    \displaystyle\arg\max_{H\in\mathcal{S}_k}\bar{P}(H), & k<K_{\mathrm{acc}}, \\[2pt]
    \displaystyle\arg\min_{H\in\mathcal{S}_k}\ell(H), & k\ge K_{\mathrm{acc}}.
  \end{cases}
  \label{eq:round}
\end{equation}
Accuracy ties favor lower latency; efficiency ties favor higher $\bar{P}$. The head is retained for equivalent
metrics or when no alternative is selected.

\endgroup
\setcounter{dbltopnumber}{1}
\section{Experiments}
\label{sec:experiments}
\begin{table*}[t]
\centering
\caption{\textbf{Performance comparison on OVO-Bench RT+BT (\%).}
Avg.\ is the question-weighted accuracy over all questions of a category (837 real-time, 631 backward-tracing), and
Overall is the accuracy over all 1{,}468 questions; prior methods are aggregated from their published per-task accuracies in the same way.
Gray rows are backbone references, \rsilabel{} denotes recursively evolved harnesses, and $^{\dagger}$ marks results of our
reimplementation.
Within each backbone, \textbf{bold} and \underline{underlined} scores mark the best and second-best methods.}
\label{tab:main}
\begingroup
\fontsize{7.5}{9}\selectfont
\setlength{\tabcolsep}{6pt}
\renewcommand{\arraystretch}{1.00}
\newcommand{\ovopending}{\phantom{00.0}}
\newcommand{\rsitag}{\hspace{0.35em}{\fontsize{6.5}{6.5}\selectfont\rsilabel}}
\newcommand{\ovocategory}[3]{\multicolumn{14}{@{}l@{}}{#1\benchdot{#2}\textit{#3}}\\[-1pt]}
\begin{tabular*}{\textwidth}{@{\extracolsep{\fill}}l*{13}{c}@{}}
\toprule
\multirow{2}{*}{\textbf{Model / Method}} & \multirow{2}{*}{\textbf{Frames}}
& \multicolumn{7}{c}{\textbf{Real-Time Visual Perception}}
& \multicolumn{4}{c}{\textbf{Backward Tracing}}
& \multirow{2}{*}{\textbf{Overall}} \\
\cmidrule(lr){3-9}\cmidrule(lr){10-13}
& & OCR & ACR & ATR & STU & FPD & OJR & \textbf{Avg.} & EPM & ASI & HLD & \textbf{Avg.} & \\
\midrule
\ovocategory{}{morseproprietary}{Proprietary pretrained models}
\modelicon{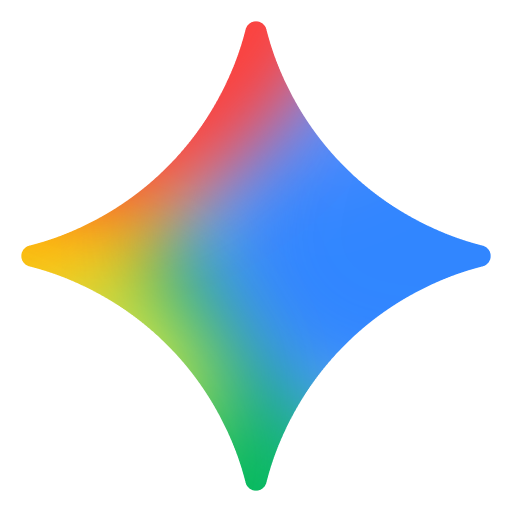}Gemini 1.5 Pro & 1\,fps & 85.9 & 67.0 & 79.3 & 58.4 & 63.4 & 62.0 & 68.7 & 58.6 & 76.4 & 52.6 & 61.0 & 65.4 \\
\modelicon{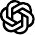}GPT-4o & 64 & 69.8 & 64.2 & 71.6 & 51.1 & 70.3 & 59.8 & 63.2 & 57.9 & 75.7 & 48.7 & 59.4 & 61.6 \\
\modelicon{openai.pdf}GPT-5-mini & 64 & 81.2 & 67.9 & 81.9 & 66.3 & 73.3 & 66.3 & 72.2 & 66.7 & 75.7 & 16.1 & 53.9 & 64.3 \\
\modelicon{openai.pdf}GPT-5.1 & 64 & 83.2 & 78.9 & 77.6 & 67.4 & 79.2 & 76.6 & 76.6 & 62.0 & 73.7 & 38.7 & 57.8 & 68.5 \\
\addlinespace[1pt]
\ovocategory{\morsegroupseparator}{morsepretrained}{Open-source pretrained models}
LLaVA-Video-7B~\cite{zhang2025llavavideo} & 64 & 69.1 & 58.7 & 68.8 & 49.4 & 74.3 & 59.8 & 62.1 & 56.2 & 57.4 & 7.5 & 42.2 & 53.5 \\
LLaVA-OneVision-7B~\cite{li2025llavaonevision} & 64 & 66.4 & 57.8 & 73.3 & 53.4 & 71.3 & 62.0 & 63.1 & 54.2 & 55.4 & 21.5 & 44.9 & 55.3 \\
LLaVA-OneVision-2-8B~\cite{an2026llavaonevision2} & 0.5\,fps & 62.4 & 62.4 & 69.0 & 61.8 & 62.0 & 54.9 & 61.5 & 53.2 & 62.8 & 21.0 & 46.0 & 54.8 \\
Qwen2-VL-7B~\cite{wang2024qwen2vl} & 64 & 60.4 & 50.5 & 56.0 & 47.2 & 66.3 & 55.4 & 55.3 & 47.8 & 35.5 & 56.1 & 47.4 & 51.9 \\
Qwen2-VL-72B~\cite{wang2024qwen2vl} & 64 & 65.8 & 60.6 & 69.8 & 51.7 & 69.3 & 54.4 & 60.6 & 52.5 & 60.8 & 57.5 & 56.0 & 58.6 \\
\addlinespace[1pt]
\ovocategory{\morsegroupseparator}{morsestreaming}{Open-source streaming models}
Flash-VStream-7B~\cite{zhang2025flashvstream} & 1\,fps & 24.2 & 29.4 & 28.5 & 33.7 & 25.7 & 28.8 & 28.7 & 39.1 & 37.2 & 5.9 & 28.8 & 28.8 \\
VideoLLM-online-8B~\cite{chen2024videollmonline} & 2\,fps & 8.1 & 23.9 & 12.1 & 14.0 & 45.5 & 21.2 & 19.4 & 22.2 & 18.8 & 12.2 & 18.5 & 19.0 \\
StreamChat~\cite{xiong2025streamchat} & 15\,fps & 51.7 & 47.7 & 58.6 & 41.0 & 61.1 & 47.8 & 50.1 & 50.2 & 54.1 & 37.6 & 47.4 & 49.0 \\
Dispider~\cite{qian2025dispider} & 1\,fps & 57.7 & 49.5 & 62.1 & 44.9 & 61.4 & 51.6 & 53.6 & 48.5 & 55.4 & 4.3 & 37.1 & 46.5 \\
\midrule
\morseref{\textbf{Qwen2.5-VL-7B}~\cite{bai2025qwen25vl}} & \morseref{0.5\,fps} & \morseref{76.5} & \morseref{57.8} & \morseref{68.1} & \morseref{46.6} & \morseref{66.3} & \morseref{56.5} & \morseref{60.9} & \morseref{49.8} & \morseref{58.1} & \morseref{44.6} & \morseref{50.2} & \morseref{56.3} \\
\hspace{0.5em}+ HERMES (4K)~\cite{zhang2026hermes} & 0.5\,fps & 85.2 & 64.2 & 71.6 & 53.4 & \underline{74.3} & 65.2 & 68.1 & 48.5 & \underline{62.2} & 37.6 & 48.5 & 59.7 \\
\hspace{0.5em}+ FluxMem~\cite{xie2026fluxmem} & 1\,fps & 81.2 & 59.6 & 70.7 & 53.4 & \textbf{75.3} & 63.0 & 66.3 & 48.5 & \textbf{64.2} & 29.0 & 46.4 & 57.8 \\
\hspace{0.5em}+ OASIS~\cite{liang2026oasis} & 48/80 & 85.2 & \underline{72.5} & 66.4 & 52.3 & 67.3 & 64.7 & 67.3 & 51.9 & 58.8 & \underline{48.9} & \underline{52.6} & 61.0 \\
\hspace{0.5em}+ StreamScout (zero-shot)~\cite{zhang2026streamscout} & 4/12/24 & \underline{92.0} & \textbf{77.1} & 74.1 & \underline{59.0} & \underline{74.3} & \underline{77.7} & \underline{75.3} & 53.5 & 60.1 & 36.6 & 50.1 & \underline{64.4} \\
\hspace{0.5em}+ MemEvo$^{\dagger}$~\cite{liu2026memevo}\rsitag & 1\,fps & \textbf{92.6} & \textbf{77.1} & \underline{79.3} & 58.4 & 73.3 & 69.6 & 74.1 & \underline{59.6} & 58.8 & 32.8 & 51.5 & \underline{64.4} \\
\morsehighlight\hspace{0.5em}\textbf{+ \method{} (Ours)}\rsitag & 1\,fps & 89.3 & 70.6 & \textbf{83.6} & \textbf{62.9} & \underline{74.3} & \textbf{81.5} & \textbf{76.9} & \textbf{59.9} & \textbf{64.2} & \textbf{58.6} & \textbf{60.5} & \textbf{69.9} \\
\midrule
\morseref{\textbf{Qwen3-VL-8B}~\cite{bai2025qwen3vl}} & \morseref{0.5\,fps} & \morseref{83.9} & \morseref{58.7} & \morseref{70.7} & \morseref{56.2} & \morseref{69.3} & \morseref{64.1} & \morseref{66.8} & \morseref{51.2} & \morseref{60.8} & \morseref{43.6} & \morseref{51.2} & \morseref{60.1} \\
\hspace{0.5em}+ FluxMem$^{\dagger}$~\cite{xie2026fluxmem} & 1\,fps & 85.9 & 71.6 & 73.3 & 57.9 & 71.3 & 66.8 & 70.4 & 58.9 & 66.2 & 22.6 & 49.9 & 61.6 \\
\hspace{0.5em}+ OASIS~\cite{liang2026oasis} & 48/80 & \underline{92.6} & 81.7 & 78.5 & 65.2 & 68.0 & 78.3 & 77.3 & 58.3 & \underline{67.6} & 51.6 & 58.5 & 69.2 \\
\hspace{0.5em}+ StreamScout (zero-shot)~\cite{zhang2026streamscout} & 4/12/24 & 91.3 & 82.6 & \textbf{82.8} & 64.0 & \underline{75.3} & 75.0 & 77.7 & 55.2 & 61.5 & \underline{63.4} & 59.1 & 69.7 \\
\hspace{0.5em}+ MemEvo$^{\dagger}$~\cite{liu2026memevo}\rsitag & 1\,fps & \underline{92.6} & \textbf{87.2} & \underline{79.3} & \underline{69.1} & 74.3 & \textbf{84.2} & \underline{81.0} & \textbf{63.0} & 66.2 & 50.5 & \underline{60.1} & \underline{72.0} \\
\morsehighlight\hspace{0.5em}\textbf{+ \method{} (Ours)}\rsitag & 1\,fps & \textbf{95.3} & \underline{84.4} & \textbf{82.8} & \textbf{71.9} & \textbf{80.2} & \underline{82.1} & \textbf{82.4} & \underline{61.6} & \textbf{73.0} & \textbf{65.1} & \textbf{65.3} & \textbf{75.1} \\
\midrule
\morseref{\textbf{Qwen3.8-27B}~\cite{qwen2026qwen38}} & \morseref{1\,fps} & \morseref{79.9} & \morseref{60.6} & \morseref{81.9} & \morseref{58.4} & \morseref{76.2} & \morseref{66.3} & \morseref{69.7} & \morseref{65.3} & \morseref{68.2} & \morseref{94.1} & \morseref{74.5} & \morseref{71.7} \\
\hspace{0.5em}+ FluxMem$^{\dagger}$~\cite{xie2026fluxmem} & 1\,fps & 90.6 & \underline{78.9} & 80.2 & \underline{70.8} & \underline{77.2} & 82.9 & \underline{80.1} & 64.7 & \underline{68.2} & \textbf{79.8} & \underline{70.0} & \underline{75.8} \\
\hspace{0.5em}+ MemEvo$^{\dagger}$~\cite{liu2026memevo}\rsitag & 1\,fps & \underline{94.0} & \textbf{85.3} & \underline{82.8} & 60.7 & \underline{77.2} & \underline{83.2} & 79.8 & \textbf{70.4} & 57.4 & 44.1 & 59.6 & 71.1 \\
\morsehighlight\hspace{0.5em}\textbf{+ \method{} (Ours)}\rsitag & 1\,fps & \textbf{94.6} & \textbf{85.3} & \textbf{85.3} & \textbf{75.8} & \textbf{83.2} & \textbf{84.8} & \textbf{84.6} & \underline{68.4} & \textbf{74.3} & \underline{77.4} & \textbf{72.4} & \textbf{79.4} \\
\bottomrule
\end{tabular*}
\endgroup
\vspace{4pt}
\end{table*}

We evaluate \method{} on three streaming video benchmarks, trace the development of the evolved harness, and
analyze its inference efficiency. We then examine the contributions of online research and tool acquisition, and
test whether the resulting harness transfers to other answering models.

\subsection{Experimental Setup}
\label{sec:exp_setup}

\noindent \textbf{Benchmarks and data.}
The \editor{} diagnoses failures on 780 training questions from 370 streams, drawn from
NExT-GQA~\cite{xiao2024nextgqa}, STAR~\cite{wu2021star}, ReXTime~\cite{chen2024rextime}, COIN~\cite{tang2019coin},
YouCook2~\cite{zhou2018youcook2}, and the COIN subset of OVBench~\cite{huang2025ovbench}. It can inspect the
training questions and execution traces. Candidate selection uses a separate validation set of 388 questions from
182 streams of the same sources; only aggregate validation scores are exposed to the \editor{}.
The two splits share no videos. For testing, we use OVO-Bench~\cite{niu2025ovobench},
StreamingBench~\cite{lin2026streamingbench}, and StreamBench~\cite{xiong2025streamchat}. The first two report
question-weighted multiple-choice accuracy, while StreamBench uses its official LLM-judged protocol for free-form
answers. Each query has access only to observations available at its timestamp. Test videos are excluded from
training, validation, and model selection, with video overlap checked across the splits.

\vspace{3pt} \noindent \textbf{Implementation.}
We evolve a separate harness for each of three frozen answering models: Qwen2.5-VL-7B~\cite{bai2025qwen25vl},
Qwen3-VL-8B-Instruct~\cite{bai2025qwen3vl}, and Qwen3.8-27B~\cite{qwen2026qwen38}, each served with vLLM and greedy
decoding. All runs sample video at 1\,fps and cap the persistent frame store at $G=64$ frames. A separate
16-second window provides recent frames, while text and embeddings have no global storage cap. Each VLM call uses at
most 32 frames and 256 visual tokens per frame. The seed $H_0$ answers from 16-second VLM captions and the four
most recent frames; these four frames remain in every answer call and do not occupy the persistent store.
The initial perception tools are CLIP ViT-L/14~\cite{radford2021clip} and OCR, alongside the answering VLM.
The \editor{} uses Claude Opus 5.5 in Claude Code with a fresh sandboxed session per round.
The broker logs all searches and page reads and blocks benchmark repositories, dataset pages, and question-answer
annotations.

\begin{table*}[t]
\centering
\caption{\textbf{Performance comparison on StreamingBench and StreamBench (\%).}
StreamingBench Avg.\ is the question-weighted accuracy over its four subsets, and StreamBench uses its official LLM-judged protocol. GPT-4o uses 64 frames for StreamingBench
and 50 for StreamBench~\cite{xiong2025streamchat}.
Gray rows are backbone references, purple \rsilabel{} tags denote recursively evolved harnesses, and $^{\dagger}$ marks results of
our reimplementation.
Within each backbone, \textbf{bold} and \underline{underlined} scores mark the best and second-best methods.}
\label{tab:stream_main}
\begingroup
\fontsize{7.5}{9}\selectfont
\setlength{\tabcolsep}{6pt}
\renewcommand{\arraystretch}{1.02}
\newcommand{\ovopending}{\phantom{00.0}}
\newcommand{\rsitag}{\hspace{0.35em}{\fontsize{6.5}{6.5}\selectfont\rsilabel}}
\newcommand{\ovocategory}[3]{\multicolumn{14}{@{}l@{}}{#1\benchdot{#2}\textit{#3}}\\[-1pt]}
\begin{tabular*}{\textwidth}{@{\extracolsep{\fill}}l*{13}{c}@{}}
\toprule
\multirow{2}{*}{\textbf{Model / Method}} & \multirow{2}{*}{\textbf{Frames}}
& \multicolumn{5}{c}{\textbf{StreamingBench}}
& \multicolumn{7}{c}{\textbf{StreamBench}} \\
\cmidrule(lr){3-7}\cmidrule(lr){8-14}
& & Real-Time & ACU & MCU & SQA & \textbf{Avg.} & OS & LM & SM & CI & KG & SF & \textbf{Avg.} \\
\midrule
\ovocategory{}{morseproprietary}{Proprietary pretrained models}
\modelicon{gemini.png}Gemini 1.5 Pro & 1\,fps & 75.7 & 51.4 & 40.7 & 54.8 & 69.5 & -- & -- & -- & -- & -- & -- & -- \\
\modelicon{openai.pdf}GPT-4o & 64 / 50 & 73.3 & 41.2 & 38.4 & 32.8 & 65.0 & 60.5 & 61.2 & 64.4 & 72.3 & 93.9 & 74.7 & 71.0 \\
\modelicon{openai.pdf}GPT-5-mini & 64 & 82.6 & 46.0 & 52.8 & 54.4 & 75.3 & 68.6 & 75.6 & 73.2 & 79.8 & 64.4 & 79.8 & 73.6 \\
\modelicon{openai.pdf}GPT-5.1 & 64 & 83.5 & 49.6 & 51.2 & 52.4 & 76.0 & 67.2 & 67.0 & 72.0 & 79.5 & 66.8 & 82.1 & 72.4 \\
\addlinespace[1pt]
\ovocategory{\morsegroupseparator}{morsestreaming}{Open-source streaming models}
Flash-VStream-7B~\cite{zhang2025flashvstream} & 1\,fps & 23.2 & -- & -- & -- & -- & 37.1 & 44.5 & 48.6 & 58.1 & 66.4 & 59.2 & 52.1 \\
VideoLLM-online-8B~\cite{chen2024videollmonline} & 5\,fps & -- & -- & -- & -- & -- & 41.4 & 48.8 & 52.9 & 62.7 & 69.2 & 64.1 & 56.4 \\
StreamChat~\cite{xiong2025streamchat} & 15\,fps & 60.1 & 28.8 & 22.4 & 32.0 & 52.6 & 40.7 & 43.6 & 47.1 & 63.0 & 89.1 & 73.8 & 59.5 \\
\midrule
\morseref{\textbf{Qwen2.5-VL-7B}~\cite{bai2025qwen25vl}} & \morseref{0.5\,fps} & \morseref{66.0} & \morseref{33.2} & \morseref{30.1} & \morseref{34.0} & \morseref{58.3} & \morseref{36.1} & \morseref{41.3} & \morseref{41.0} & \morseref{50.5} & \morseref{76.5} & \morseref{62.3} & \morseref{51.1} \\
\hspace{0.5em}+ OASIS~\cite{liang2026oasis} & 48/80 & 70.6 & 38.8 & 36.0 & 46.0 & 63.6 & \underline{35.9} & 39.4 & 44.1 & \textbf{54.5} & 76.9 & 64.9 & \underline{52.4} \\
\hspace{0.5em}+ StreamScout (zero-shot)~\cite{zhang2026streamscout} & 4/12/24 & 77.7 & \textbf{58.0} & 39.2 & \underline{53.2} & 71.3 & 34.8 & 32.0 & 43.5 & 41.4 & \underline{83.6} & 63.8 & 49.7 \\
\hspace{0.5em}+ MemEvo$^{\dagger}$~\cite{liu2026memevo}\rsitag & 1\,fps & \underline{78.3} & 54.0 & \underline{50.0} & 47.6 & \underline{71.9} & 33.8 & \underline{42.2} & \underline{49.6} & 47.5 & 55.0 & \underline{68.2} & 49.4 \\
\morsehighlight\hspace{0.5em}\textbf{+ \method{} (Ours)}\rsitag & 1\,fps & \textbf{82.3} & \underline{56.0} & \textbf{50.4} & \textbf{56.4} & \textbf{75.8} & \textbf{42.8} & \textbf{43.9} & \textbf{57.2} & \underline{50.8} & \textbf{86.2} & \textbf{76.2} & \textbf{59.5} \\
\midrule
\morseref{\textbf{Qwen3-VL-8B}~\cite{bai2025qwen3vl}} & \morseref{0.5\,fps} & \morseref{72.8} & \morseref{35.6} & \morseref{35.7} & \morseref{43.9} & \morseref{64.9} & \morseref{33.8} & \morseref{50.5} & \morseref{47.8} & \morseref{69.4} & \morseref{73.5} & \morseref{62.6} & \morseref{56.0} \\
\hspace{0.5em}+ OASIS~\cite{liang2026oasis} & 48/80 & 78.2 & 42.7 & 49.6 & 48.4 & 71.0 & 47.0 & 53.6 & 54.3 & 71.7 & \textbf{86.1} & 67.6 & 62.1 \\
\hspace{0.5em}+ StreamScout (zero-shot)~\cite{zhang2026streamscout} & 4/12/24 & 80.3 & 48.0 & 42.0 & 42.0 & 71.9 & 52.5 & 51.8 & 61.4 & 71.4 & \underline{84.6} & 70.8 & \underline{65.3} \\
\hspace{0.5em}+ MemEvo$^{\dagger}$~\cite{liu2026memevo}\rsitag & 1\,fps & \underline{81.8} & \underline{57.2} & \underline{59.6} & \underline{56.4} & \underline{76.2} & \underline{57.5} & \underline{55.1} & \underline{65.2} & \textbf{76.1} & 26.5 & \underline{81.5} & 60.5 \\
\morsehighlight\hspace{0.5em}\textbf{+ \method{} (Ours)}\rsitag & 1\,fps & \textbf{84.2} & \textbf{61.2} & \textbf{64.4} & \textbf{58.4} & \textbf{78.9} & \textbf{65.9} & \textbf{57.4} & \textbf{73.5} & \underline{73.7} & 70.8 & \textbf{88.1} & \textbf{71.6} \\
\midrule
\morseref{\textbf{Qwen3.8-27B}~\cite{qwen2026qwen38}} & \morseref{1\,fps} & \morseref{80.1} & \morseref{44.8} & \morseref{34.0} & \morseref{43.6} & \morseref{71.0} & \morseref{44.8} & \morseref{58.1} & \morseref{57.2} & \morseref{76.1} & \morseref{75.8} & \morseref{73.5} & \morseref{64.1} \\
\hspace{0.5em}+ MemEvo$^{\dagger}$~\cite{liu2026memevo}\rsitag & 1\,fps & \underline{78.0} & \underline{44.0} & \underline{56.4} & \underline{41.6} & \underline{70.9} & \textbf{82.6} & \underline{65.4} & \textbf{79.7} & \underline{79.1} & \underline{40.6} & \textbf{86.1} & \underline{72.4} \\
\morsehighlight\hspace{0.5em}\textbf{+ \method{} (Ours)}\rsitag & 1\,fps & \textbf{89.0} & \textbf{73.2} & \textbf{71.6} & \textbf{70.0} & \textbf{85.0} & \underline{76.3} & \textbf{72.0} & \underline{79.1} & \textbf{81.5} & \textbf{73.8} & \underline{85.1} & \textbf{78.0} \\
\bottomrule
\end{tabular*}
\endgroup
\end{table*}

\vspace{3pt} \noindent \textbf{Evolution settings.}
We run $K_{\mathrm{acc}}=20$ accuracy rounds followed by $K_{\mathrm{eff}}=5$ efficiency rounds. Each round proposes
at most $B=6$ initial candidates before merges, with a research budget $B_{\mathrm{res}}$ of 30 paper requests and
40 web requests. Accuracy updates must increase the mean correct-option probability by at least $\delta=0.004$.
Efficiency updates use $\rho=0.9$ to require at least a 10\% reduction in answer latency, while losing at most
$\epsilon=0.003$ in mean correct-option probability and $\kappa=2$ correct answers relative to the current harness.
Total losses relative to the start of the efficiency stage are capped at $\epsilon_{\mathrm{tot}}=0.008$ and
$\kappa_{\mathrm{tot}}=4$ correct answers. During evolution, both stages apply the $L=1$\,s gate to estimated mean
latency on the validation set and enforce $\omega(H)\le1$, using fixed call-cost coefficients to avoid server-load
fluctuations. Test-time latency and write cost are measured on a single NVIDIA H100 GPU.

\subsection{Main Results}
\label{sec:exp_main}

\noindent \textbf{Results on OVO-Bench.}
\method{} achieves the highest overall accuracy within all three backbone groups in \cref{tab:main}.
With Qwen3-VL-8B, it reaches 75.1\%, ahead of the manually designed StreamScout at 69.7\% and the evolved
MemEvo program at 72.0\%. The task breakdown helps explain this advantage. Relative to MemEvo, the margin is
larger on backward tracing than on real-time perception, indicating that better use of past observations is an
important source of improvement. The same pattern holds with Qwen2.5-VL-7B, where backward-tracing accuracy
reaches 60.5\%, 7.9 points above the best competing score on this subset. With Qwen3.8-27B, the larger advantage
over FluxMem, the strongest competing harness overall, is instead in real-time perception. Harness evolution benefits different model
sizes and temporal tasks, although the capabilities that improve most differ across backbones.

\vspace{3pt} \noindent \textbf{Results on StreamingBench and StreamBench.}
As shown in \cref{tab:stream_main}, \method{} achieves the highest average score within all three backbone groups
on both benchmarks. On StreamingBench, Qwen3-VL-8B improves all four subsets over MemEvo. The gains on anomaly
and misleading context understanding suggest that the evolved harness supplies useful historical evidence for
interpreting current observations. On StreamBench, the same backbone reaches 71.6\%, compared with 65.3\% for
StreamScout, the strongest competing harness. It leads in four of six categories, with clear advantages in object
search and short-term memory search. These results support the usefulness of the evolved memory and retrieval
policies for both contextual reasoning and free-form answers, beyond the multiple-choice format used during
evolution. With this backbone, other harnesses remain stronger on conversational interaction and knowledge-based
questions.

The advantage also holds in a domain absent from evolution: on the driving videos of
ODV-Bench~\cite{zeng2025streamforest}, \method{} reaches 67.1\% with Qwen3-VL-8B, compared with 62.6\% for MemEvo
and 58.6\% for the model alone (\cref{app:results}).

\subsection{Evolution Analysis}
\label{sec:exp_evolution}
\label{sec:exp_analysis}
\begin{figure*}[t]
  \centering
  \includegraphics[width=\linewidth]{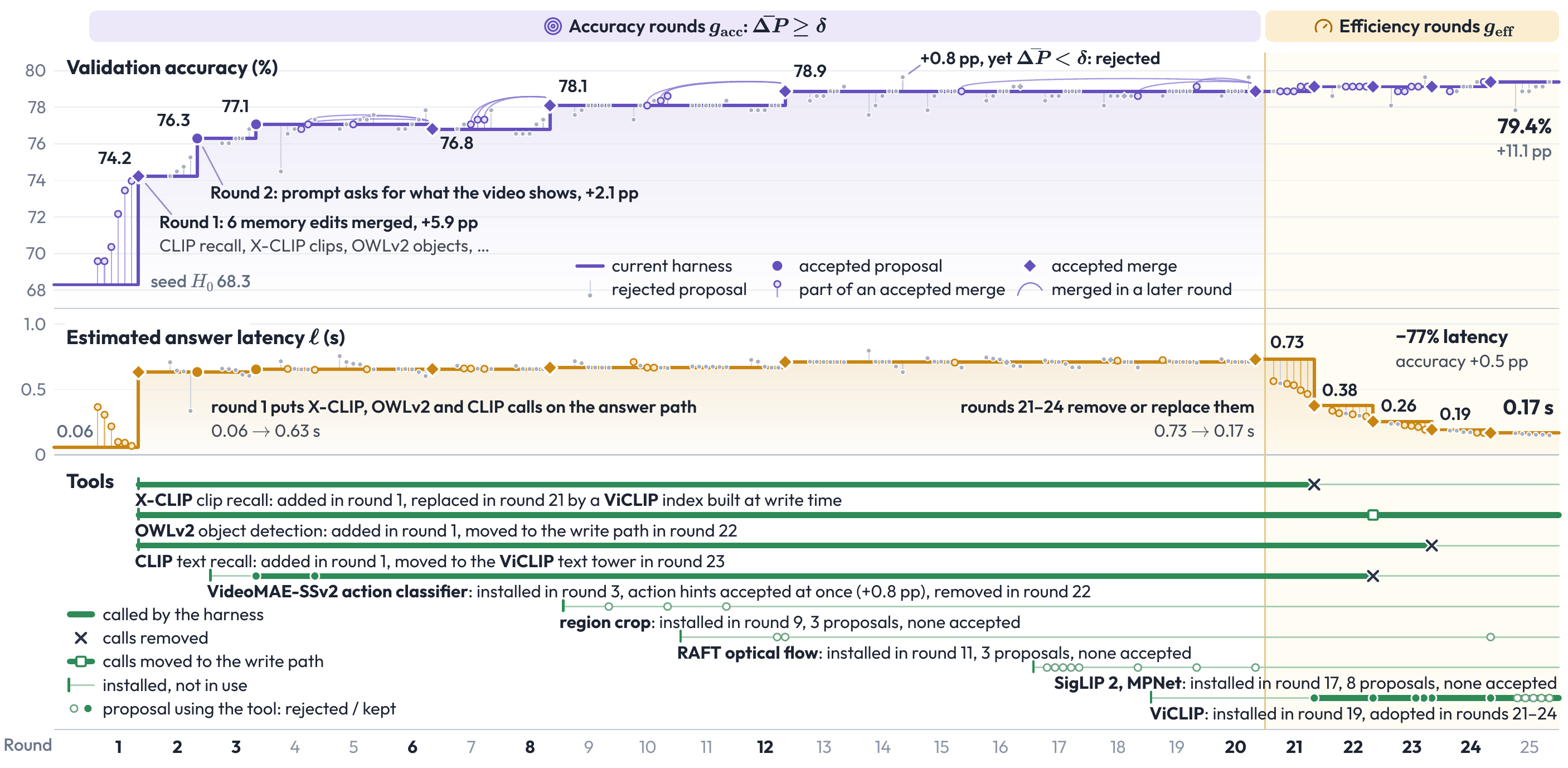}
  \caption{\textbf{Harness evolution with Qwen3-VL-8B.} \emph{Top \& middle:} accuracy and estimated answer latency on
  the validation set. Lines track the current harness, while small marks show evaluated candidates. Large circles and diamonds denote
  accepted individual revisions.
  \emph{Bottom:} tool availability and use. Thick segments indicate use by the accepted harness; crosses mark
  removal and open squares mark relocation to the write path. Circles indicate candidates using a newly installed
  tool, filled when they enter an accepted harness. \looseness=-1}
  \label{fig:evolution}
\end{figure*}

\label{sec:exp_tools}
In \cref{fig:evolution}, the Qwen3-VL-8B harness first improves evidence preservation and retrieval, then
simplifies execution. Missing past evidence motivates visual recall, while event-order errors motivate chapter
summaries and a step log. X-CLIP~\cite{ni2022xclip}, OWLv2~\cite{minderer2023owlv2}, and
VideoMAE~\cite{tong2022videomae} provide clip matching, object selection, and action cues.

\begin{table}[t]
\centering
\caption{\textbf{Test accuracy along the evolution (\%).} Qwen3-VL-8B checkpoints: the
seed $H_0$, $H_{\mathrm{S3}}$ after round~3, $H_{\mathrm{S7}}$ after the accuracy stage, and the final
harness $H_{\mathrm{S11}}$. RT and BT are the real-time and backward-tracing averages; Gain is the improvement of
the final harness over the seed.}
\label{tab:checkpoints}
\begingroup
\footnotesize
\setlength{\tabcolsep}{0pt}
\setlength{\morsenumw}{0.128\columnwidth}
\renewcommand{\arraystretch}{1.14}
\newcommand{\cksub}[1]{{\fontsize{6.4}{6.4}\selectfont\color{morsereference}#1}}
\begin{tabular*}{\columnwidth}{@{}l@{\hspace{0.9em}}l@{\extracolsep{\fill}}wc{\morsenumw}@{\extracolsep{0pt}}wc{\morsenumw}wc{\morsenumw}wc{\morsenumw}@{\hspace{0.02\columnwidth}}wc{0.125\columnwidth}@{}}
\toprule
\multicolumn{2}{@{}l}{\textbf{Benchmark}} & $H_0$ & $H_{\mathrm{S3}}$ & $H_{\mathrm{S7}}$ & $H_{\mathrm{S11}}$ & \textbf{Gain} \\[-2.5pt]
 & & \cksub{seed} & \cksub{round 3} & \cksub{round 20} & \cksub{round 24} & \cksub{over seed} \\
\midrule
OVO-Bench & RT & 78.4 & 80.0 & 82.7 & \cellcolor{morseoursband}82.4 & \textcolor{morseyes}{+4.0} \\
 & BT & 52.8 & 61.3 & 64.8 & \cellcolor{morseoursband}65.3 & \textcolor{morseyes}{+12.5} \\
 & Overall & 67.4 & 72.0 & 75.0 & \cellcolor{morseoursband}75.1 & \textcolor{morseyes}{+7.7} \\
\multicolumn{2}{@{}l}{StreamingBench} & 72.8 & 77.2 & 79.3 & \cellcolor{morseoursband}78.9 & \textcolor{morseyes}{+6.1} \\
\multicolumn{2}{@{}l}{StreamBench} & 64.1 & 68.0 & 70.8 & \cellcolor{morseoursband}71.6 & \textcolor{morseyes}{+7.5} \\
\bottomrule
\end{tabular*}
\endgroup
\end{table}

Validation keeps tool use selective: cropping and optical flow are explored without entering the accepted
harness. Meanwhile, the arcs trace code from rejected candidates that the \editor{} reapplies to the current head
and submits as new combinations. Validation tests each combined program, allowing the search to build on earlier attempts.
\Cref{tab:checkpoints} evaluates four checkpoints chosen by their positions in the run: the seed $H_0$,
$H_{\mathrm{S3}}$ after round~3, $H_{\mathrm{S7}}$ after the accuracy stage, and the final $H_{\mathrm{S11}}$,
the eleventh accepted update. By $H_{\mathrm{S3}}$, past-frame recall, a revised prompt, and an action classifier
produce the largest jump in backward tracing. Later accuracy rounds further improve all three test benchmarks,
extending progress beyond the initial recall mechanism. Overall, backward-tracing accuracy rises from 52.8\% to
65.3\%, against 78.4\% to 82.4\% in real-time perception. Since the seed already supplies recent frames, evolution
brings its larger benefit to questions about the past.

The efficiency stage moves object detection to memory construction, replaces repeated clip scoring with a
write-time ViCLIP~\cite{wang2023internvid} index, and removes action-classifier calls. Originally requested for
procedure-step boundaries, ViCLIP instead enables cheaper retrieval, showing how an acquired capability can
serve a new purpose. From $H_{\mathrm{S7}}$ to $H_{\mathrm{S11}}$, estimated validation latency falls from 0.73\,s to
0.17\,s, while test accuracy changes by at most 0.8 points per benchmark. Reorganizing computation thus preserves
the accuracy-stage gains with less query-time work. \Cref{app:reading} details the research and revision history.

\subsection{Efficiency Analysis}
\label{sec:exp_efficiency}

\begin{figure}[t]
  \centering
  \includegraphics[width=\linewidth]{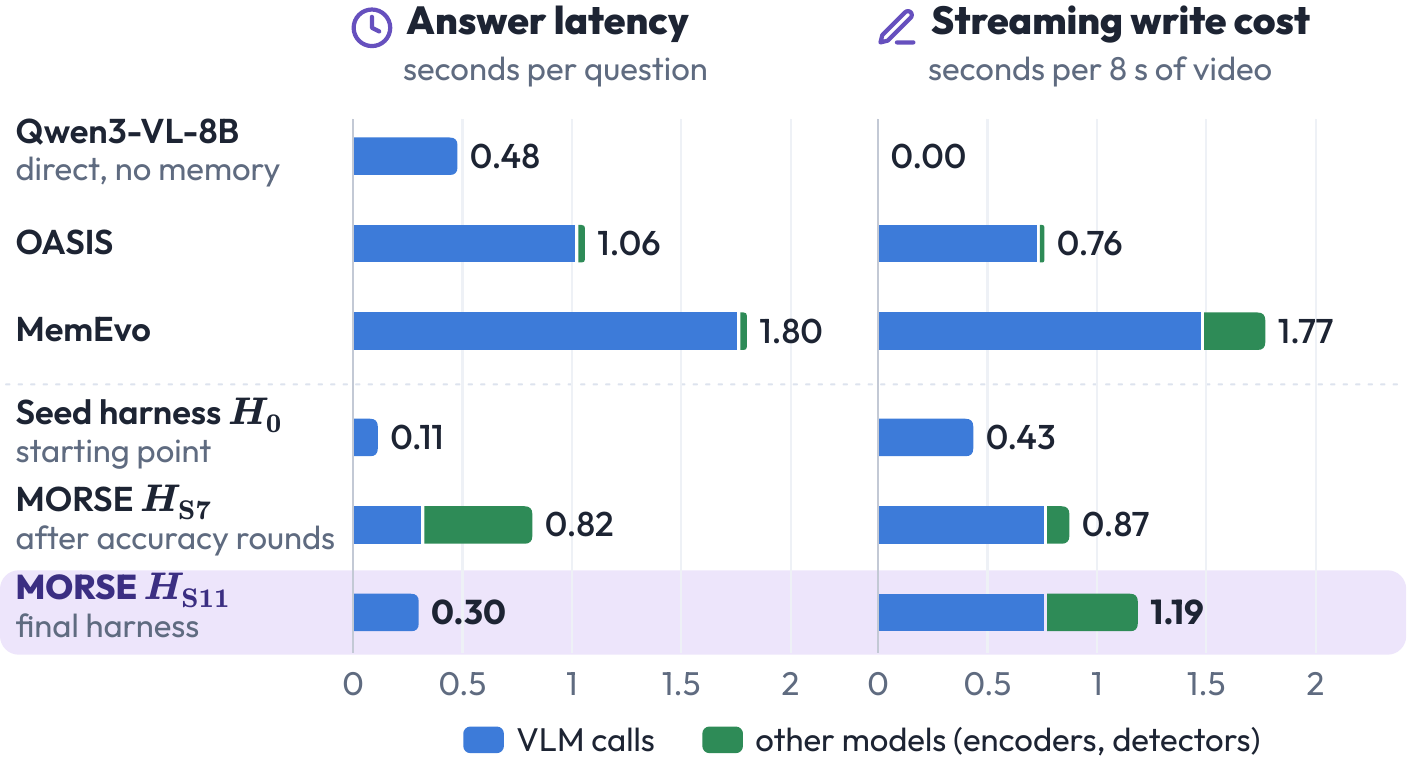}
  \caption{\textbf{Inference efficiency on OVO-Bench.} Measured mean answer latency and streaming write cost,
  using Qwen3-VL-8B for all methods. Each bar separates VLM calls from calls to other
  models. The direct VLM uses at most 128 frames at 1\,fps without memory; $H_0$, $H_{\mathrm{S7}}$, and
  $H_{\mathrm{S11}}$ are the seed, the harness after the accuracy rounds, and the final harness.}
  \label{fig:efficiency}
\end{figure}

Streaming inference incurs both continuous memory-construction cost and the latency experienced when a question
arrives. \Cref{fig:efficiency} separates these costs on OVO-Bench using Qwen3-VL-8B for every method.
\method{} answers in 0.30\,s on average, 6.0$\times$ faster than MemEvo and 3.5$\times$ faster than OASIS.
Together with the higher OVO-Bench accuracy in \cref{tab:main}, this shows that richer historical evidence need
not require slower answers. It is also 1.6$\times$ faster than answering directly from sampled frames.
The seed $H_0$ remains faster, but cannot retrieve past frames once they leave the recent window, as reflected
in the backward-tracing results in \cref{tab:checkpoints}.
The cost breakdown shows where this speed comes from. After the accuracy rounds, $H_{\mathrm{S7}}$ answers in
0.82\,s, of which 0.49\,s goes to encoder and detector calls made when a question arrives. The efficiency rounds
move reusable computation to memory construction and remove redundant calls: the write cost rises from
0.87\,s to 1.19\,s per 8\,s of video, and answer
latency falls to 0.30\,s, making the final harness 2.7$\times$ faster than $H_{\mathrm{S7}}$. Its write cost is
still 33\% below MemEvo's, because VLM calls consume 0.77\,s against MemEvo's 1.49\,s, a saving that more than
offsets the encoder and detector calls. OASIS spends only 0.76\,s on writing but answers more slowly and less
accurately.
\Cref{app:results} gives further cost and token results.

\begin{table}[t]
\centering
\caption{\textbf{Contribution of the open-world components on OVO-Bench (\%).} The seed $H_0$ is evaluated before
evolution; each of the other three variants runs 25 rounds from this seed with Qwen3-VL-8B. RT and
BT are the real-time and backward-tracing averages; subscripts give the gain over the seed harness $H_0$.}
\label{tab:ablation}
\begingroup
\footnotesize
\setlength{\tabcolsep}{0pt}
\setlength{\morsenumw}{0.135\columnwidth}
\renewcommand{\arraystretch}{1.14}
\begin{tabular*}{\columnwidth}{@{}l@{\extracolsep{\fill}}wc{0.15\columnwidth}@{\extracolsep{0pt}}wc{0.1\columnwidth}@{\hspace{0.03\columnwidth}}wc{\morsenumw}wc{\morsenumw}wc{\morsenumw}@{}}
\toprule
\textbf{Variant} & \textbf{Research} & \textbf{Tools} & \thd{\textbf{RT}} & \thd{\textbf{BT}} & \thd{\textbf{Overall}} \\
\midrule
\morseref{Seed harness $H_0$} & \morseref{--} & \morseref{--} & \morseref{\tv{78.4}} & \morseref{\tv{52.8}} & \morseref{\tv{67.4}} \\
w/o tool requests & \yesmark & \nomark & \tup{78.9}{0.5} & \tup{59.7}{6.9} & \tup{70.6}{3.2} \\
w/o online research & \nomark & \yesmark & \tup{80.8}{2.4} & \tup{62.4}{9.6} & \tup{72.9}{5.5} \\
\morsehighlightcol\textbf{\method{} (Ours)} & \yesmark & \yesmark & \tup{82.4}{4.0} & \tup{65.3}{12.5} & \tup{75.1}{7.7} \\
\bottomrule
\end{tabular*}
\endgroup
\end{table}

\subsection{Further Discussions}
\label{sec:exp_discussion}

\noindent \textbf{Contribution of different components.}
\label{sec:exp_ablation}
We separate the contribution of consulting external resources from that of expanding the available tools.
Starting from the same seed $H_0$, we repeat evolution with Qwen3-VL-8B under the same data splits, round and
candidate budgets, and gates. Without online research, the \editor{} can still request tools from its prior
knowledge, subject to host review. Without tool requests, it can still read papers, model cards, and
code, but must build every candidate from the initial inventory $\mathcal{T}_0$.

As shown in \cref{tab:ablation}, removing online research lowers OVO-Bench accuracy from 75.1\% to 72.9\% even
though tool acquisition remains available. This supports the value of consulting existing work beyond what the
\editor{} can propose from its own knowledge. Conversely, online research with a fixed tool inventory reaches
70.6\%. Program revision within this inventory still improves the seed's 67.4\%, but the full method benefits
further from making new perception capabilities available. Both removals hurt backward tracing more than real-time
perception, linking their contribution particularly to the preservation and retrieval of historical evidence.
These comparisons support using online research and tool acquisition together in the evolution loop.
\par

\begin{table}[t]
\centering
\caption{\textbf{Transfer to other answering models on OVO-Bench (\%).} Each model is evaluated alone, with
MemEvo, and with the harness that \method{} evolved with Qwen3-VL-8B, applied unchanged with its tools. Subscripts
give the change relative to the model alone. RT and BT are the real-time and backward-tracing averages, and
$^{\dagger}$ marks results of our reimplementation.}
\label{tab:transfer}
\begingroup
\footnotesize
\setlength{\tabcolsep}{0pt}
\setlength{\morsenumw}{0.16\columnwidth}
\renewcommand{\arraystretch}{1.14}
\begin{tabular*}{\columnwidth}{@{}l@{\extracolsep{\fill}}wc{\morsenumw}@{\extracolsep{0pt}}wc{\morsenumw}wc{\morsenumw}@{}}
\toprule
\textbf{Model / Method} & \thd{\textbf{RT}} & \thd{\textbf{BT}} & \thd{\textbf{Overall}} \\
\midrule
\morseref{\textbf{LLaVA-OneVision-2-8B}~\cite{an2026llavaonevision2}} & \morseref{\tv{61.5}} & \morseref{\tv{46.0}} & \morseref{\tv{54.8}} \\
\hspace{0.5em}+ MemEvo$^{\dagger}$~\cite{liu2026memevo} & \tup{74.8}{13.3} & \tup{52.8}{6.8} & \tup{65.3}{10.5} \\
\morsehighlightcol\hspace{0.5em}\textbf{+ \method{} (Ours)} & \tup{81.0}{19.5} & \tup{58.0}{12.0} & \tup{71.1}{16.3} \\
\midrule
\morseref{\textbf{Gemma-4-26B-A4B}~\cite{google2026gemma4}} & \morseref{\tv{71.7}} & \morseref{\tv{57.1}} & \morseref{\tv{65.4}} \\
\hspace{0.5em}+ MemEvo$^{\dagger}$~\cite{liu2026memevo} & \tup{77.5}{5.8} & \tup{62.0}{4.9} & \tup{70.8}{5.4} \\
\morsehighlightcol\hspace{0.5em}\textbf{+ \method{} (Ours)} & \tup{82.6}{10.9} & \tup{65.1}{8.0} & \tup{75.1}{9.7} \\
\bottomrule
\end{tabular*}
\endgroup
\end{table}

\vspace{3pt} \noindent \textbf{Transfer across answering models.}
\label{sec:exp_transfer}
The main results use a separately evolved harness for each backbone. Here we test whether the revisions remain
useful to models whose failures did not guide evolution. We keep the program and tools of the Qwen3-VL-8B
harness $H_{\mathrm{S11}}$ unchanged and replace the answering model with
LLaVA-OneVision-2-8B~\cite{an2026llavaonevision2} or Gemma-4-26B-A4B~\cite{google2026gemma4}, two models from
other families. \Cref{tab:transfer} compares the transferred harness with each target model alone and with
MemEvo applied to the same model.
With LLaVA-OneVision-2-8B, the transferred harness raises OVO-Bench accuracy from 54.8\% to 71.1\% without
another evolution run, improving real-time perception by 19.5 points and backward tracing by 12.0 points. With
Gemma-4-26B-A4B, it raises accuracy from 65.4\% to 75.1\%. MemEvo also improves both models, to 65.3\% and
70.8\%, but \method{} stays 5.8 and 4.3 points ahead. This
suggests that the revised evidence organization remains useful across answering models.

\setcounter{dbltopnumber}{2}
\section{Conclusion}
\label{sec:conclusion}

We presented \method{}, an open-world auto-research framework that uses recurring failures to guide online
research for improving streaming video harnesses. Its central idea is to seek out and adapt existing solutions,
allowing memory and retrieval policies to evolve together with their perception tools while the answering
VLM remains frozen. Experiments on OVO-Bench, StreamingBench, and StreamBench show improvements over competing
harnesses across answering models, and the OVO-Bench efficiency evaluation shows lower answer latency than
MemEvo and OASIS. The evolution analysis illustrates how capabilities introduced to address failures can later
be reorganized to reduce computation. These findings support a practical approach to recursive self-improvement:
learning when and how to reuse published methods and models, with validation determining which changes benefit
the harness.

\section*{Acknowledgments}
This work has been funded in part by the ARL award W911QX-24-F-0049, DARPA award FA8750-23-2-1015, ONR award N00014-23-1-2840, ONR MURI grant N00014-25-1-2116, and USDA-NIFA award 2021-67021-35329.

\par
{
    \small
    \bibliographystyle{ieeenat_fullname}
    \bibliography{main}
}

\clearpage
\hypersetup{pageanchor=false}
\backrefsetup{disable}
\setcounter{page}{1}
\appendix
\maketitlesupplementary

\section*{Overview}
This supplementary material describes the implementation and evaluation of \method{}, documents the research
and revisions behind the evolved harness, and provides detailed results and memory examples. It is organized
into four appendices:
\begin{itemize}[leftmargin=*,topsep=4pt,itemsep=4pt,partopsep=0pt]
  \item \textbf{Appendix~\ref{app:implementation}: Implementation and Evaluation Protocol.}
  We specify the host interfaces, tool-admission rules, runtime and storage limits, scoring and candidate
  selection, and the data splits and overlap checks used during evolution and evaluation.
  \item \textbf{Appendix~\ref{app:reading}: Research and Harness Evolution.}
  We trace the \editor{}'s literature research across rounds, connect tool requests to the failures that motivated
  them, summarize the eleven accepted harness updates and their validation results, and examine how accepted updates
  affect the failures they target and the remaining questions.
  \item \textbf{Appendix~\ref{app:results}: Additional Quantitative Results.}
  We report latency, write-cost, and prompt-token results, give StreamingBench results by task and
  backbone, and describe our reproductions of MemEvo, OASIS, and the direct VLM baseline.
  \item \textbf{Appendix~\ref{app:memory_examples}: Memory at Question Time.}
  Three OVO-Bench examples illustrate what the final harness stores, retrieves, and sends to the answering VLM
  for episodic recall, action ordering, and future prediction.
\end{itemize}
Throughout this material, \emph{the main run} is the evolution run with Qwen3-VL-8B analyzed in
the main paper, and $H_{\mathrm{S11}}$ is its final harness.

\section{Implementation and Evaluation Protocol}
\label{app:implementation}
\subsection{Host Protocol and Runtime Limits}
\label{sec:suppl_cost}
\label{app:protocol}
\vspace{3pt} \noindent \textbf{Research access.}
The \editor{} uses a host-side broker for paper search and reading, searches over released models and code, and
page reads from approved documentation hosts. The broker logs requests, enforces the per-round budget
$B_{\mathrm{res}}$, and blocks sources of training, validation, and test data, including benchmark repositories,
dataset pages, question-answer annotations, and other answer-bearing files. Retrieved material informs candidate development without exposing evaluation
content, a concern also identified in recursive video-harness optimization~\cite{lin2026videoharnesssrsi}.

\vspace{3pt} \noindent \textbf{Tool requests and admission.}
Candidates cannot install executable content from the network or load additional models directly. A tool request
specifies the model and its interface, the training failures it targets, and how its outputs would enter the harness.
It includes the expected per-call latency, license, training-data provenance, and links to the paper and model page.
The host excludes models trained or fine-tuned on evaluation videos or annotations and admits only non-generative
models, keeping the frozen VLM as the answer generator. An approved service becomes available after the host deploys
its backend and registers its interface and timing profile.
Requests can remain pending across rounds. Every call counts toward the answer latency or the write load in
\cref{eq:cost}.

\vspace{3pt} \noindent \textbf{Storage and context.}
The host enforces a persistent store of 64 frames and a separate 16-second readable window. The four most recent
frames required in each answer come from this window and do not use persistent slots. Each VLM call can contain at
most 32 frames, leaving at most 28 additional frames in $H_{\mathrm{S11}}$. Cropped images cannot be persisted.
Text and vectors have no host-wide storage cap: host captions and the final harness's segment embeddings can grow
with the stream. Other structures use program-defined limits, including 120 step-span notes, 40 chapter notes,
and 48 per-frame CLIP vectors in $H_{\mathrm{S11}}$. The host also limits each episode to 20,000 service calls and
each call to 131,072 context tokens.

\subsection{Scoring and Candidate Selection}
\label{app:selection}
\vspace{3pt} \noindent \textbf{Option probabilities.}
The scorer finds the last generated token that normalizes to a valid option letter, stripping surrounding spaces
and the punctuation \texttt{.():*}. At that position, it aggregates the returned top-20 token log-probabilities
$\lambda_i(v)$ over spelling variants of each valid option $a$:
\begin{equation}
  \begin{aligned}
    m_i(a)&=\sum_{\substack{v\in\operatorname{Top20}_i\\\operatorname{norm}(v)=a}}
      \exp\lambda_i(v), \\
    p_i&=\frac{m_i(y_i)}{\sum_{a\in\mathcal{O}_i}m_i(a)},
  \end{aligned}
\end{equation}
where $\mathcal{O}_i$ is the set of valid letters. A correct option absent from the returned alternatives receives
zero mass. If several answer calls record probabilities, the last recorded distribution is used. An answered
question without a recorded distribution falls back to its binary correctness; unanswered questions receive zero.

\vspace{3pt} \noindent \textbf{Latency during evolution.}
The gates compare a candidate with its parent on the same streams, so the latency they use should not vary between
repeated runs of the same program. During evolution, we therefore estimate the time of each service call with a
linear function of its input size, such as the number of frames, texts, and prompt and completion tokens, fitted
once and kept fixed. Summing these estimates over the calls of a question or a stream gives the answer latency and
write load used by the gates, which are the values shown for each round in \cref{fig:evolution} and
\cref{tab:updates}. These estimates guide candidate screening and selection during evolution. All latency and
write-cost results reported for the test benchmarks, including \cref{fig:efficiency}, are measured on a single
NVIDIA H100 GPU.

\vspace{3pt} \noindent \textbf{Selection and history.}
The main run maintains a single head. Accuracy selection forms Pareto layers using higher $\bar{P}$ and lower
latency, treating latency differences within 2\% as equivalent, and takes the largest $\bar{P}$ on the first layer.
Points differing by less than $10^{-4}$ in $\bar{P}$ and within 2\% in latency are collapsed in pool order:
the head first, then current candidates in proposal order, their merges, and finally archived candidates.
Efficiency selection excludes the archive and orders eligible current candidates by increasing latency, breaking
ties by higher $\bar{P}$. Training results
inform diagnosis but do not determine acceptance.

Each round starts a new \editor{} session; diagnosis, planning, editing, and merging within that round share the
session. The workspace supplies current code, per-question training traces and histories, aggregate validation
scores, previous analysis and edit summaries, candidate diffs, reusable observations, and service admission status.
The host retains snapshots of candidates that passed the gate without being selected, for archive selection. The
\editor{} can also apply saved diffs from rejected candidates to a copy of the current head and submit the
combination as a new proposal. This is how rounds 6, 8, 12, and 20 reused earlier candidates. The combined programs
underwent engineering checks and validation, passed the gates, and had the highest $\bar{P}$ in their selection
pools. The driver's separate merge stage combines only
current-round candidates that pass the gate and share a parent, then revalidates the resulting programs.

\subsection{Evaluation Subsets and Data Separation}
\label{app:data}
Evolution uses training and validation sets constructed from NExT-GQA~\cite{xiao2024nextgqa}, STAR~\cite{wu2021star},
ReXTime~\cite{chen2024rextime}, COIN~\cite{tang2019coin}, YouCook2~\cite{zhou2018youcook2}, and the COIN subset of
OVBench~\cite{huang2025ovbench}. The training split contains 780 questions on 370 streams, and validation contains
388 questions on 182 streams.

OVO-Bench evaluation includes six real-time perception tasks (OCR, ACR, ATR, STU, FPD, and OJR) and three
backward-tracing tasks (EPM, ASI, and HLD), totaling 1{,}468 questions on 512 videos. We exclude forward active
responding. StreamingBench evaluation includes
real-time visual understanding, anomalous and misleading contextual understanding, and sequential question answering. Omni-source and proactive-output
tasks are excluded. Sequential questions use the official history.

Training and validation share no videos, and neither split shares a video with OVO-Bench,
StreamingBench, or StreamBench; overlap is checked with video IDs, source links, and frame fingerprints.

\section{Research and Harness Evolution}
\label{app:reading}
\subsection{External Research}
The research broker records every request of the \editor{}, and the analysis report of each round states what the
\editor{} concluded from the material it read. This section reconstructs the main run from these records. The counts
quoted below are the \editor{}'s own measurements on training questions.

\vspace{3pt} \noindent \textbf{Volume.}
Over the 25 rounds the broker served 153 requests: 79 literature searches, 53 paper reads, 13 searches of model and
code hubs, and 8 page fetches (six model cards or code repositories and two arXiv abstracts).
The searches returned 587 results covering 209 distinct papers, and the \editor{} opened 28 of them
(\cref{tab:reading_list}), in 22 of the 25 rounds. Counting the two abstracts, 50 of the 55 paper reads fall in the
accuracy stage. Several
papers were consulted repeatedly: VideoAgent~\cite{fan2024videoagent} and R4DSG~\cite{ma2026r4dsg} six times each,
in five rounds each. Eighteen of the 28 papers appeared in 2026 and seven in 2025.

\begin{table}[t]
\centering
\caption{\textbf{Papers opened by the \editor{} in the main run}, with the rounds in which each was read.}
\label{tab:reading_list}
\footnotesize
\setlength{\tabcolsep}{4pt}
\begin{tabular}{@{}lr@{}}
\toprule
Paper & Rounds \\
\midrule
\multicolumn{2}{@{}l}{\emph{Streaming-video memory}} \\
Vista~\cite{lu2026vista} & 1 \\
OASIS~\cite{liang2026oasis} & 3 \\
EventMemAgent~\cite{wen2026eventmemagent} & 3 \\
Online textual memory~\cite{cali2026textmemory} & 3, 5, 11, 17 \\
Memory Helps, but Confabulation Misleads~\cite{zhang2025confabulation} & 5, 12, 17 \\
StreamMeCo~\cite{wang2026streammeco} & 5 \\
Event-VStream~\cite{guo2026eventvstream} & 6, 10 \\
SVMemAgent~\cite{ko2026svmemagent} & 7, 19 \\
Think While Watching~\cite{wang2026thinkwatch} & 8 \\
Visual Agentic Memory~\cite{li2026vam} & 11 \\
ProV-Mem~\cite{wang2026provmem} & 14 \\
SPACE~\cite{niu2026space} & 19 \\
TimeChat-Online~\cite{yao2025timechatonline} & 24 \\
\addlinespace[2pt]
\multicolumn{2}{@{}l}{\emph{Long-video memory and agents}} \\
VideoAgent~\cite{fan2024videoagent} & 4, 5, 6, 11, 17 \\
WorldMM~\cite{yeo2025worldmm} & 4 \\
R4DSG~\cite{ma2026r4dsg} & 5, 11, 12, 14, 18 \\
Video-EM~\cite{wang2025videoem} & 8, 11 \\
From Verbatim to Gist~\cite{lian2026mmmem} & 11 \\
Glance and Focus~\cite{bai2024glancefocus} & 12 \\
\addlinespace[2pt]
\multicolumn{2}{@{}l}{\emph{Keyframe and token selection}} \\
Coverage-driven keyframe selection~\cite{zhang2026cses} & 3, 13, 20, 22 \\
Divide, then Ground~\cite{li2025dig} & 8 \\
Adaptive Keyframe Sampling~\cite{tang2025aks} & 9, 15 \\
ReQuest~\cite{kim2026request} & 14, 15 \\
QCA~\cite{peng2026qca} & 20 \\
KTV~\cite{song2026ktv} & 22 \\
\addlinespace[2pt]
\multicolumn{2}{@{}l}{\emph{Perception aids and encoders}} \\
MLLMs Know Where to Look~\cite{zhang2025vicrop} & 9 \\
HiERO-StepG~\cite{zenotto2026hierostepg} & 16 \\
InternVid~\cite{wang2023internvid} & 18, 23 \\
\bottomrule
\end{tabular}
\end{table}

\vspace{3pt} \noindent \textbf{Rounds 1--3: recovering past evidence, and the first tool.}
The seed answers from captions and the four most recent frames. In round~1 the \editor{} re-asked the 209 training
questions that the seed answered incorrectly with 12 uniformly sampled past frames added: 45 became correct, and for
32 of these the past frames, not the added reasoning step, made the difference. It read Vista~\cite{lu2026vista} and
proposed six ways to retain and recall past frames, among them CLIP recall by question, X-CLIP~\cite{ni2022xclip}
clip scoring, and OWLv2~\cite{minderer2023owlv2} object frames. All six passed the gate, and their merge raised
accuracy by 5.9 points, the largest step of the run. In round~2 it found that 46 incorrect questions about brief
hand--object interactions are decided by one to three frames and that the VLM's choice flips under a neutral
rewording in 5 of 23 cases. It searched a model hub, located a VideoMAE classifier fine-tuned on
Something-Something-v2~\cite{tong2022videomae}, read its model card, and requested it. The classifier was installed
in round~3, and the action hints built on it passed the gate (0.8 points). In between, round~2 accepted a
one-sentence change to the answer prompt (2.1 points). The round-3 report also concluded that most remaining errors
were not decision errors: the VLM ``does not see the gold event in its frames,'' and about three quarters of the
confident wrong answers were attributed to a perception or label-granularity limit. This diagnosis directed the
following rounds toward memory and perception.

\vspace{3pt} \noindent \textbf{Rounds 3--7: text memory.}
From round~3 the \editor{} read memory designs for streaming and long video: coverage-driven keyframe
selection~\cite{zhang2026cses}, online textual memory~\cite{cali2026textmemory}, OASIS~\cite{liang2026oasis},
EventMemAgent~\cite{wen2026eventmemagent}, VideoAgent~\cite{fan2024videoagent}, WorldMM~\cite{yeo2025worldmm},
R4DSG~\cite{ma2026r4dsg}, StreamMeCo~\cite{wang2026streammeco}, Event-VStream~\cite{guo2026eventvstream}, and
SVMemAgent~\cite{ko2026svmemagent}. Informed by these designs, it implemented several text memories: an event log,
a step log, chapter summaries, and verified event entries. The literature also helped it interpret what it observed.
A write-time action log missed the deciding interaction on six of six inspected videos and wrote confident wrong
lines, which the \editor{} related to the confabulation effect reported in~\cite{zhang2025confabulation}. An
always-on event log fixed 24 training questions but broke 21, the distractor effect of retrieving more text reported
in~\cite{cali2026textmemory}. From these it derived two rules for the rounds that followed: to ``write few,
verified, specific entries'' and to ``retrieve less.'' Designs that need services the host does not provide, such as
the tracking and segmentation of R4DSG, or that exceed the latency and write budgets, such as the multi-turn
retrieval of WorldMM, were set aside. Chapter summaries entered the accepted harness in round~6 as part of a merge.

\vspace{3pt} \noindent \textbf{Rounds 8--11: perception aids.}
In round~8 the \editor{} found that, for hand--object questions, the asked object was not visible to the VLM at the
\solver{}'s frame size in five of eight inspected wrong cases, against one of five correct controls. It read Divide,
then Ground~\cite{li2025dig}, Think While Watching~\cite{wang2026thinkwatch}, and Video-EM~\cite{wang2025videoem},
concluded that text entries need faithfulness checks rather than more entries, and requested a hand--object contact
detector~\cite{shan2020handobject} and a region-crop operation. Region crop was installed in round~9. In that round
the \editor{} read a study of visual cropping~\cite{zhang2025vicrop}, which reports that accuracy depends on the size
of the visual subject and that sending a crop with the full image helps, and proposed detector-guided crops that
follow this recipe. In round~10 it observed that the scene often stays unchanged when a step ends, so whole-frame
change points miss the end, and requested RAFT optical flow~\cite{teed2020raft} as a hand-region motion signal,
which was installed in round~11. These two services were explored without entering the accepted harness
(\cref{tab:tool_requests}).

\begin{table*}[t]
\centering
\caption{\textbf{Service requests after round 1.} Reasons are condensed from the requests; fractions are the
\editor{}'s measurements on training questions.}
\label{tab:tool_requests}
\footnotesize
\setlength{\tabcolsep}{5pt}
\begin{tabular}{@{}cp{0.21\textwidth}p{0.40\textwidth}p{0.27\textwidth}@{}}
\toprule
Round & Requested service & Reason given & Outcome \\
\midrule
2 & VideoMAE classifier fine-tuned on Something-Something-v2~\cite{tong2022videomae} &
Hand--object interactions are decided by one to three frames, and the VLM's choice is unstable under rewording
(5 of 23). & Installed in round 3; adopted in round 3; calls removed in round 22. \\
\addlinespace[2pt]
8 & Hand--object contact detector~\cite{shan2020handobject} &
The asked object is not visible to the VLM at the \solver{}'s frame size (5 of 8 wrong cases); contact events could
be logged at write time. & Pending at the end of the run. \\
\addlinespace[2pt]
8 & Region crop (host image operation) &
Asked objects are too small at the \solver{}'s frame size; a crop of the detector box shows them larger. &
Installed in round 9; explored by three candidates, not adopted. \\
\addlinespace[2pt]
10 & RAFT optical flow~\cite{teed2020raft} &
The scene stays unchanged when a step ends, so whole-frame change points miss the end; motion in the hand region may
mark it. & Installed in round 11; explored by three candidates, not adopted. \\
\addlinespace[2pt]
16 & Sentence encoder (all-mpnet-base-v2)~\cite{reimers2019sbert} &
The CLIP text encoder is weak at matching logged step lines to option phrases (7 of 17). &
Installed in round 17; explored, not adopted. \\
\addlinespace[2pt]
16 & SigLIP~2~\cite{tschannen2025siglip2} &
Locating a named step on the per-second CLIP image trace is weak (24 of 43). &
Installed in round 17; explored, not adopted. \\
\addlinespace[2pt]
18 & ViCLIP-L/14~\cite{wang2023internvid} &
Locating where the anchor step ends is the bottleneck for before/after questions (36 of 89). &
Installed in round 19; adopted in rounds 21--24. \\
\addlinespace[2pt]
23 & Batched CLIP text endpoint~\cite{radford2021clip} &
CLIP text embedding is 22\% of the answer latency and grows with the number of texts. & Not installed. \\
\bottomrule
\end{tabular}
\end{table*}

\vspace{3pt} \noindent \textbf{Rounds 10--15: a fine-grained step log.}
An anchor-based step-log hint was proposed in round~10 but fell below the probability gate. In round~11, reading
VideoAgent, Video-EM, Visual Agentic Memory~\cite{li2026vam}, MM-Mem~\cite{lian2026mmmem}, online textual
memory~\cite{cali2026textmemory}, and R4DSG informed further experiments. Training counterexamples showed that
the hint could recommend the wrong option or be ignored. The \editor{} therefore tested a short excerpt of the
anchor and its next two log entries in place of an explicit option recommendation, a choice also supported by
Video-EM and online textual memory. Trigger and anchor adjustments came from training-set calibration. Neither
this excerpt variant nor a separate duration-oriented ledger passed the gate. The combination accepted in
round~12 reused the round-10 step-log code unchanged. Round~12
read Glance and Focus~\cite{bai2024glancefocus} as the closest published design to anchor localization, and the
confabulation study motivated a candidate that checks caption claims against the detector. In rounds 13--15 the
\editor{} read keyframe-selection and streaming-memory methods, including Adaptive Keyframe
Sampling~\cite{tang2025aks}, ReQuest~\cite{kim2026request}, and ProV-Mem~\cite{wang2026provmem}, and tested their
premise on its own traces: $\bar{P}$ is flat in the share of held evidence that is sent (0.766, 0.754, and 0.777).
Frame selection was therefore not the limiting factor, and the following rounds turned to step localization.

\vspace{3pt} \noindent \textbf{Rounds 16--20: encoders for step localization.}
For questions about what happens before or after a named step, the \editor{} measured that matching logged step lines
to the options with the CLIP text encoder names the correct option in only 7 of 17 cases in which the log contains the
answer span. Round~16 read HiERO-StepG~\cite{zenotto2026hierostepg} and requested a sentence
encoder~\cite{reimers2019sbert} and SigLIP~2~\cite{tschannen2025siglip2}, both installed in round~17. Round~17 found that the text log repeats the VLM's own misperceptions and concluded that only a
non-generative signal could correct them, which is what the acquired perception models provide. Round~18 identified
locating the end of the anchor step as the bottleneck (the anchor was right in 36 of 89 cases), searched a model hub
for video-text encoders, read the InternVid paper~\cite{wang2023internvid}, and requested ViCLIP. ViCLIP was installed in round~19 and later became part of the accepted harness. Rounds 19 and 20
read SVMemAgent, SPACE~\cite{niu2026space}, and QCA~\cite{peng2026qca}; the update accepted in round~20 merged three
earlier frame-coverage candidates.

\vspace{3pt} \noindent \textbf{Rounds 21--25: efficiency.}
In the efficiency stage, research served cost reduction. A write-time ViCLIP index replaced answer-time X-CLIP
scoring in round~21. The InternVid paper states that both ViCLIP encoders are initialized from CLIP, which suggested
the ViCLIP text tower as a cheaper substitute for CLIP text embeddings: the fixed cost model assigns 3.5\,ms per
call plus 0.5\,ms per text, compared with 36\,ms per text for CLIP. The revision accepted in round~23 routes all text embeddings there after a
label-free check that the two encoders give nearly identical similarities. On the 30 training questions for which
ViCLIP was requested, estimated mean answer latency falls from 0.95\,s under $H_{\mathrm{S7}}$ to 0.28\,s under the final harness. KTV~\cite{song2026ktv} and
TimeChat-Online~\cite{yao2025timechatonline} reduce visual tokens by exploiting the redundancy of video. Because the
host fixes the tokens per frame, the \editor{} reduced redundancy across frames instead, and the revision accepted
in round~24 removes near-duplicate key frames.

\vspace{3pt} \noindent \textbf{How research shaped the harness.}
The record links research to the accepted harness in three ways. First, it supplied mechanisms: the recall of past
frames accepted in round~1 followed the reading of Vista, and the chapter summaries accepted in round~6 followed
the earlier research on text memory. Second, it supplied models: the action
classifier and ViCLIP were located through model hubs and papers after a specific failure had been measured, and
each request in \cref{tab:tool_requests} cites that failure together with a model card, a repository, or a paper.
Third, it supplied principles and facts that guided later edits, such as writing few, verified entries, and the
shared initialization of the ViCLIP and CLIP encoders that enabled the text-encoder substitution of round~23.
Validation then decided which of these ideas stayed in the harness.

\subsection{Tool Requests and Adoption}
\label{app:tools}
\Cref{tab:tool_requests} lists the eight service requests after the first-round X-CLIP and OWLv2 acquisition, with
the reason given by the \editor{} and the outcome. Six of these services were installed, and two of them, the action classifier and ViCLIP, entered the
accepted harness. The hand--object detector was still pending at the end of the run, and the batched text endpoint
was not installed.

\begin{table*}[t]
\centering
\caption{\textbf{The eleven accepted updates of the main run.} Correct is out of the 388 validation questions,
$\bar{P}$ is the mean probability of the correct option, and latency is the estimated mean answer latency on the
validation set. In rounds 6, 8, 12, and 20, the \editor{} reapplied earlier diffs to the current head and submitted
each combination as a new candidate. Rounds 21--24 are efficiency rounds.}
\label{tab:updates}
\footnotesize
\setlength{\tabcolsep}{5pt}
\begin{tabular}{@{}clp{0.56\textwidth}ccc@{}}
\toprule
Round & Harness & Accepted change & Correct & $\bar{P}$ & \shortstack{Estimated\\latency (s)} \\
\midrule
-- & $H_0$ & Seed: all captions and the four most recent frames. & 265 & 0.678 & 0.06 \\
\addlinespace[2pt]
1 & $H_{\mathrm{S1}}$ & Merge of six candidates that retain and recall past frames: recency coverage, CLIP recall by
question, X-CLIP clip scoring, frames of the recent span without a caption, OWLv2 object frames, and a step timeline.
& 288 & 0.740 & 0.63 \\
\addlinespace[2pt]
2 & $H_{\mathrm{S2}}$ & One sentence in the answer prompt that prefers the option stating what the video shows. &
296 & 0.760 & 0.63 \\
\addlinespace[2pt]
3 & $H_{\mathrm{S3}}$ & Action hints from the VideoMAE classifier for hand--object questions with short options. &
299 & 0.765 & 0.65 \\
\addlinespace[2pt]
6 & $H_{\mathrm{S4}}$ & Reapplied diffs from three candidates in rounds 4 and 5: verb-gated action hints, a query for the start
of a step, and chapter summaries. & 298 & 0.769 & 0.66 \\
\addlinespace[2pt]
8 & $H_{\mathrm{S5}}$ & Reapplied diffs from three round-7 candidates, with a trigger adjustment: a check for ``still performing'' questions, step tags
for questions about the last $N$ seconds, and CLIP recall for each option. & 303 & 0.774 & 0.67 \\
\addlinespace[2pt]
12 & $H_{\mathrm{S6}}$ & Reapplied diffs from three round-10 candidates: novelty-aware eviction in the frame store, a
write-time step log read through an anchor-located hint, and the placement of the duration note. & 306 & 0.783 &
0.71 \\
\addlinespace[2pt]
20 & $H_{\mathrm{S7}}$ & Reapplied diffs from rounds 15, 18, and 19 that set frame coverage by question
time: captions and recent frames only up to 6\,s, every readable frame up to 31\,s, and gap-filling key frames beyond
64\,s. & 306 & 0.789 & 0.73 \\
\midrule
21 & $H_{\mathrm{S8}}$ & Merge of five candidates: discarded and duplicate work removed, a narrower OWLv2 trigger,
fewer X-CLIP clips, a write-time ViCLIP index in place of answer-time X-CLIP, and fewer CLIP recall texts. & 307 &
0.789 & 0.38 \\
\addlinespace[2pt]
22 & $H_{\mathrm{S9}}$ & Merge of four candidates: OWLv2 detection moved to the write path, the answer-time OWLv2
and VideoMAE calls removed, a ViCLIP window index for recall, and one CLIP recall text instead of one per option. &
307 & 0.786 & 0.26 \\
\addlinespace[2pt]
23 & $H_{\mathrm{S10}}$ & Merge of four candidates: text embeddings moved to the ViCLIP text encoder, and a narrower
scope for the step-log hint. & 307 & 0.785 & 0.19 \\
\addlinespace[2pt]
24 & $H_{\mathrm{S11}}$ & Merge of two candidates that drop near-duplicate key frames. & 308 & 0.785 & 0.17 \\
\bottomrule
\end{tabular}
\end{table*}

\subsection{Accepted Updates of the Main Run}
\label{app:updates}
\Cref{tab:updates} lists the eleven updates behind the steps of \cref{fig:evolution}. The run produced 155
candidates: 150 \editor{} proposals, including six near-miss combinations, and five additional candidates from
the merge stage.
In the accuracy stage, 13 of 121 proposed
candidates passed the accuracy gate and 7 of the 20 rounds accepted an update. Four of these seven updates (rounds 6,
8, 12, and 20) combine saved diffs from earlier rejected candidates as new proposals. In the efficiency stage, 19 of 34 proposed
candidates passed the efficiency gate and four of the five rounds accepted an update.

The first three updates add recall of past frames, one sentence in the answer prompt, and the action classifier;
together they account for 8.8 of the 10.6 points gained in the accuracy stage. Two of them, the recall of past
frames and the action classifier, follow from external research: the reading of Vista in round~1 and the model
search of round~2. Each of the four later accuracy
updates combines three smaller changes to the notes and to frame coverage. The four efficiency updates reduce
estimated validation latency from 0.73\,s to 0.17\,s, while the number of correct answers rises from 306 to 308.

\subsection{From Diagnosis to Outcome}
\label{app:diagnosis_outcome}
A revision is proposed for a specific group of failures, but it is accepted on validation questions of all types.
We examine whether accepted updates correct their declared training targets and how they affect the remaining
questions.

\vspace{3pt} \noindent \textbf{A tool addresses a diagnosed failure type.}
In round~2 the \editor{} requested an action classifier after diagnosing 46 STAR training questions that
$H_{\mathrm{S1}}$ answered incorrectly. These failures concern brief hand--object interactions decided by one to
three frames. The harness accepted in round~3, $H_{\mathrm{S3}}$, invokes the classifier on 43 of these 46
questions. Relative to its parent $H_{\mathrm{S2}}$, it corrects 6 questions without breaking any in the full
46-question set and raises the mean correct-option probability over that set by 0.097. By comparison, the
preceding prompt update from $H_{\mathrm{S1}}$ to $H_{\mathrm{S2}}$ corrected only 1 of these 46 questions.

\begin{table}[!htb]
\centering
\caption{\textbf{Corrected and broken questions for the accepted accuracy updates.} Targets are the training
questions an update declares it addresses (for a merge, the union over the merged candidates); the parent answers
them incorrectly, so they can only be corrected. Each row compares the accepted harness with its parent.
Other training questions and the disjoint set of 388 validation questions are given as corrected / broken.}
\label{tab:diag_accepted}
\footnotesize
\setlength{\tabcolsep}{4.5pt}
\begin{tabular}{@{}clcccc@{}}
\toprule
 & & \multicolumn{2}{c}{Targets} & Other & \\
\cmidrule(lr){3-4}
Round & Harness & all & corrected & training & Validation \\
\midrule
1 & $H_{\mathrm{S1}}$ & 117 & 37 & 17 / 31 & 33 / 10 \\
2 & $H_{\mathrm{S2}}$ & 31 & 7 & 12 / 13 & 13 / 5 \\
3 & $H_{\mathrm{S3}}$ & 21 & 6 & 1 / 7 & 5 / 2 \\
6 & $H_{\mathrm{S4}}$ & 17 & 9 & 9 / 4 & 5 / 6 \\
8 & $H_{\mathrm{S5}}$ & 9 & 5 & 0 / 8 & 5 / 0 \\
12 & $H_{\mathrm{S6}}$ & 14 & 1 & 5 / 8 & 5 / 2 \\
20 & $H_{\mathrm{S7}}$ & 9 & 9 & 0 / 7 & 2 / 2 \\
\bottomrule
\end{tabular}
\end{table}

\vspace{3pt} \noindent \textbf{The same update moves other questions in both directions.}
Outside its targets, the round-3 update corrects 1 training question and breaks 7, so its net effect on the
training set is zero even though the targeted failures improve. \Cref{tab:diag_accepted} shows the same pattern for
every accepted accuracy update: each corrects part of its targets, and each also changes questions it was not
designed for, breaking between 4 and 31 of them. Evidence added for one type of question can mislead the VLM on
others, the distractor effect noted in the research trace.

\vspace{3pt} \noindent \textbf{Why the loop validates, merges, and diagnoses again.}
These side effects are the reason \method{} does not accept a revision on the failures it targets. Acceptance is
decided on held-out validation questions of all types, where the seven accuracy updates correct 68 questions and
break 27 in total. Candidates that help different question types are merged so that their gains combine, and every
round diagnoses the current harness anew, so the questions an update breaks are examined again in later rounds.

\section{Additional Quantitative Results}
\label{app:results}
This appendix gives detailed inference costs, a per-task view of StreamingBench, and the settings of the reproduced
baselines.

\subsection{Inference Cost in Detail}
\label{app:cost}
\Cref{tab:efficiency} gives the numbers behind \cref{fig:efficiency}, with the 90th-percentile latency, the prompt
tokens of the answer path, and the harness before the efficiency stage ($H_{\mathrm{S7}}$).

\begin{table}[!htb]
\centering
\caption{\textbf{Inference efficiency on OVO-Bench.} All systems use Qwen3-VL-8B. Write and latency summarize
memory construction and answering costs; tokens count
answer-path VLM prompt tokens per question. $H_{\mathrm{S7}}$ and $H_{\mathrm{S11}}$ are the harnesses before and after
the efficiency stage. The direct VLM answers from frames sampled at 1\,fps (at most 128, 448\,px) and builds no
memory.}
\label{tab:efficiency}
\footnotesize
\setlength{\tabcolsep}{2.3pt}
\begin{tabular*}{\columnwidth}{@{\extracolsep{\fill}}lcccc@{}}
\toprule
\multirow{2}{*}{Method} & Write & \multicolumn{2}{c}{Latency (s)} & Tokens \\
\cmidrule(lr){3-4}
 & (s / 8\,s) & mean & p90 & (k) \\
\midrule
Qwen3-VL-8B, direct~\cite{bai2025qwen3vl} & 0.00 & 0.48 & 0.63 & 14.2 \\
MemEvo~\cite{liu2026memevo} & 1.77 & 1.80 & 2.48 & 26.9 \\
OASIS~\cite{liang2026oasis} & 0.76 & 1.06 & 1.49 & 11.4 \\
\midrule
\method{} ($H_{\mathrm{S7}}$) & 0.87 & 0.82 & 0.97 & 9.6 \\
\method{} ($H_{\mathrm{S11}}$) & 1.19 & 0.30 & 0.39 & 8.8 \\
\bottomrule
\end{tabular*}
\end{table}

\vspace{3pt} \noindent \textbf{Other benchmarks.}
Mean answer latency is 0.30--0.33\,s on the three StreamingBench groups and rises to 1.14\,s on StreamBench,
where answers are free-form and require more output tokens. Its
write load, the write-path seconds per second of video, averages 0.14 to 0.17 across benchmarks and stays below 0.19
on every video, where 1 corresponds to real time.

\subsection{StreamingBench Real-Time Tasks}
\label{app:sb_tasks}
\Cref{tab:sb_tasks} breaks the real-time column of \cref{tab:stream_main} down by task for MemEvo and \method{}.
\method{} is higher on 27 of the 30 pairs of task and backbone. The largest margins are on counting (10.1, 13.3, and
20.7 points on the three backbones) and, with Qwen3.8-27B, on causal reasoning (29.7 points) and clip summarization
(22.4 points), where MemEvo falls well below its own results with the smaller backbones. MemEvo is ahead on
text-rich understanding with Qwen2.5-VL-7B and Qwen3-VL-8B (by 0.6 and 0.7 points) and on attribute perception with
Qwen3-VL-8B (by 1.0 point). Counting remains the hardest task for both methods.

\begin{table}[!htb]
\centering
\caption{\textbf{StreamingBench real-time understanding by task} for MemEvo and \method{}
(Ours). OP: object perception; CR: causal reasoning; CS: clip summarization; ATP: attribute perception; EU: event
understanding; TR: text-rich understanding; PR: prospective reasoning; SU: spatial understanding; ACP: action
perception; CT: counting. Bold marks the higher value of each pair.}
\label{tab:sb_tasks}
\footnotesize
\setlength{\tabcolsep}{3.4pt}
\begin{tabular}{@{}lcccccc@{}}
\toprule
 & \multicolumn{2}{c}{Qwen2.5-VL-7B} & \multicolumn{2}{c}{Qwen3-VL-8B} & \multicolumn{2}{c}{Qwen3.8-27B} \\
\cmidrule(lr){2-3}\cmidrule(lr){4-5}\cmidrule(l){6-7}
Task & MemEvo & Ours & MemEvo & Ours & MemEvo & Ours \\
\midrule
OP & 83.4 & \textbf{87.0} & 86.7 & \textbf{88.9} & 86.7 & \textbf{91.6} \\
CR & 72.7 & \textbf{73.4} & 71.9 & \textbf{78.1} & 61.7 & \textbf{91.4} \\
CS & 88.3 & \textbf{94.0} & 92.1 & \textbf{92.4} & 72.6 & \textbf{95.0} \\
ATP & 87.2 & \textbf{89.4} & \textbf{92.3} & 91.3 & 92.6 & \textbf{94.2} \\
EU & 75.9 & \textbf{78.5} & 81.0 & \textbf{82.3} & 75.3 & \textbf{87.3} \\
TR & \textbf{90.0} & 89.4 & \textbf{91.9} & 91.2 & 90.0 & \textbf{93.1} \\
PR & 79.6 & \textbf{88.9} & 84.3 & \textbf{88.0} & 75.0 & \textbf{85.2} \\
SU & 73.6 & \textbf{77.6} & 76.8 & \textbf{80.1} & 76.0 & \textbf{81.7} \\
ACP & 71.1 & \textbf{76.9} & 77.7 & \textbf{80.6} & 74.6 & \textbf{88.9} \\
CT & 41.0 & \textbf{51.1} & 39.9 & \textbf{53.2} & 49.5 & \textbf{70.2} \\
\midrule
All & 78.3 & \textbf{82.3} & 81.8 & \textbf{84.2} & 78.0 & \textbf{89.0} \\
\bottomrule
\end{tabular}
\end{table}

\subsection{ODV-Bench}
\label{app:odv}
ODV-Bench~\cite{zeng2025streamforest} asks 6{,}348 multiple-choice questions on 1{,}190 driving videos of at most
90\,s, each at its own time stamp with only the video up to that point available. Driving is not among the sources
of our training and validation questions. We stream the videos at 1\,fps and answer with Qwen3-VL-8B in all systems
of \cref{tab:odv}. The VLM alone, given the frames up to the question, reaches 58.6\%. \method{} reaches 67.1\%,
8.5 points above it and 4.5 points above MemEvo (62.6\%), and is ahead of MemEvo on two of the three categories.
The largest margins over MemEvo are on key information extraction (15.1 points), risk prediction (14.7), location
prediction (13.8), and risk analysis (7.2). MemEvo is ahead on questions about single static objects (72.7\% against
70.8\%), mainly on hallucination detection (7.3 points) and past traffic memory (3.0).

\begin{table}[!htb]
\centering
\caption{\textbf{ODV-Bench by task} (accuracy, \%), with Qwen3-VL-8B as the answering model throughout.
Direct: the VLM alone on at most 128 frames at 1\,fps; $^{\dagger}$: our reimplementation. The three groups are the
benchmark's categories. Bold marks the best value of each row.}
\label{tab:odv}
\footnotesize
\setlength{\tabcolsep}{4pt}
\begin{tabular}{@{}lrccc@{}}
\toprule
Task & $n$ & Direct & MemEvo$^{\dagger}$ & Ours \\
\midrule
\multicolumn{5}{@{}l}{\textit{Single static objects}} \\
\hspace{0.5em}Driving decision-making & 303 & 41.6 & \textbf{49.8} & 47.5 \\
\hspace{0.5em}Hallucination detection & 123 & 44.7 & \textbf{88.6} & 81.3 \\
\hspace{0.5em}Key information extraction & 53 & 58.5 & 66.0 & \textbf{81.1} \\
\hspace{0.5em}Past traffic memory & 574 & 62.7 & \textbf{76.3} & 73.3 \\
\hspace{0.5em}Real-time traffic perception & 531 & 62.9 & \textbf{78.5} & 77.6 \\
\hspace{0.5em}Traffic change detection & 55 & 72.7 & \textbf{74.5} & \textbf{74.5} \\
\hspace{0.5em}\textit{Group} & 1639 & 57.7 & \textbf{72.7} & 70.8 \\
\midrule
\multicolumn{5}{@{}l}{\textit{Single dynamic objects}} \\
\hspace{0.5em}Action prediction & 694 & 49.1 & 61.4 & \textbf{62.4} \\
\hspace{0.5em}Distance prediction & 1488 & 47.3 & 47.0 & \textbf{48.1} \\
\hspace{0.5em}Location prediction & 817 & 80.0 & 67.8 & \textbf{81.6} \\
\hspace{0.5em}\textit{Group} & 2999 & 56.7 & 56.0 & \textbf{60.6} \\
\midrule
\multicolumn{5}{@{}l}{\textit{Multi-object scenarios and events}} \\
\hspace{0.5em}Accident reason answering & 156 & 59.0 & \textbf{70.5} & 69.2 \\
\hspace{0.5em}Risk analysis & 625 & 81.8 & 79.4 & \textbf{86.6} \\
\hspace{0.5em}Risk prediction & 929 & 50.6 & 53.5 & \textbf{68.2} \\
\hspace{0.5em}\textit{Group} & 1710 & 62.7 & 64.5 & \textbf{75.0} \\
\midrule
All & 6348 & 58.6 & 62.6 & \textbf{67.1} \\
\bottomrule
\end{tabular}
\end{table}

\subsection{Baseline Reproduction}
\label{app:baselines}
\vspace{3pt} \noindent \textbf{FluxMem.}
The Qwen3-VL-8B and Qwen3.8-27B results are our reproductions of FluxMem~\cite{xie2026fluxmem}, sampled at 1\,fps
with the corresponding frozen backbone. The Qwen2.5-VL-7B row retains the published result.

\vspace{3pt} \noindent \textbf{MemEvo.}
We evaluate the final program published by MemEvo~\cite{liu2026memevo} rather than repeating its search, which
relies on a proprietary model. The program keeps the four most recent frames, an episodic store of at most 64 frames,
and a gist memory. The episodic store admits a frame after a histogram change or a 16-s gap and evicts by coverage
and motion. The gist memory summarizes each 8-s segment as text with pooled visual tokens and merges its oldest
adjacent entries. A question reads the recent frames, the eight best-matching episodic frames, and the four
best-matching gist entries. The backbone serves as writer and reader, retrieval keys come from Qwen3-VL-Embedding-2B,
and, for multiple-choice questions, the answer is the highest-scoring option letter of one forward pass. Where the paper leaves a choice open, we
measure the histogram change as the Bhattacharyya distance between $8{\times}8{\times}8$ HSV histograms against a
causal running median, compute Farneback flow on 160-pixel gray frames, and prompt the writer for a segment caption
of at most 128 tokens. The visual gist of a segment is a
$2{\times}2$ mosaic of four of its frames encoded with 128 visual tokens.

\vspace{3pt} \noindent \textbf{OASIS and the direct VLM.}
For the efficiency comparison, we run the released OASIS pipeline~\cite{liang2026oasis} with its official prompts on
the same Qwen3-VL-8B server, with Qwen3-Embedding-0.6B for retrieval. The direct VLM receives
the frames from the start of the video to the question at 1\,fps, thinned to at most 128 frames at 448 pixels, and
answers in one call.

\section{Memory at Question Time}
\label{app:memory_examples}
\begin{figure*}[p]
\centering
\includegraphics[width=\textwidth,height=0.85\textheight,keepaspectratio]{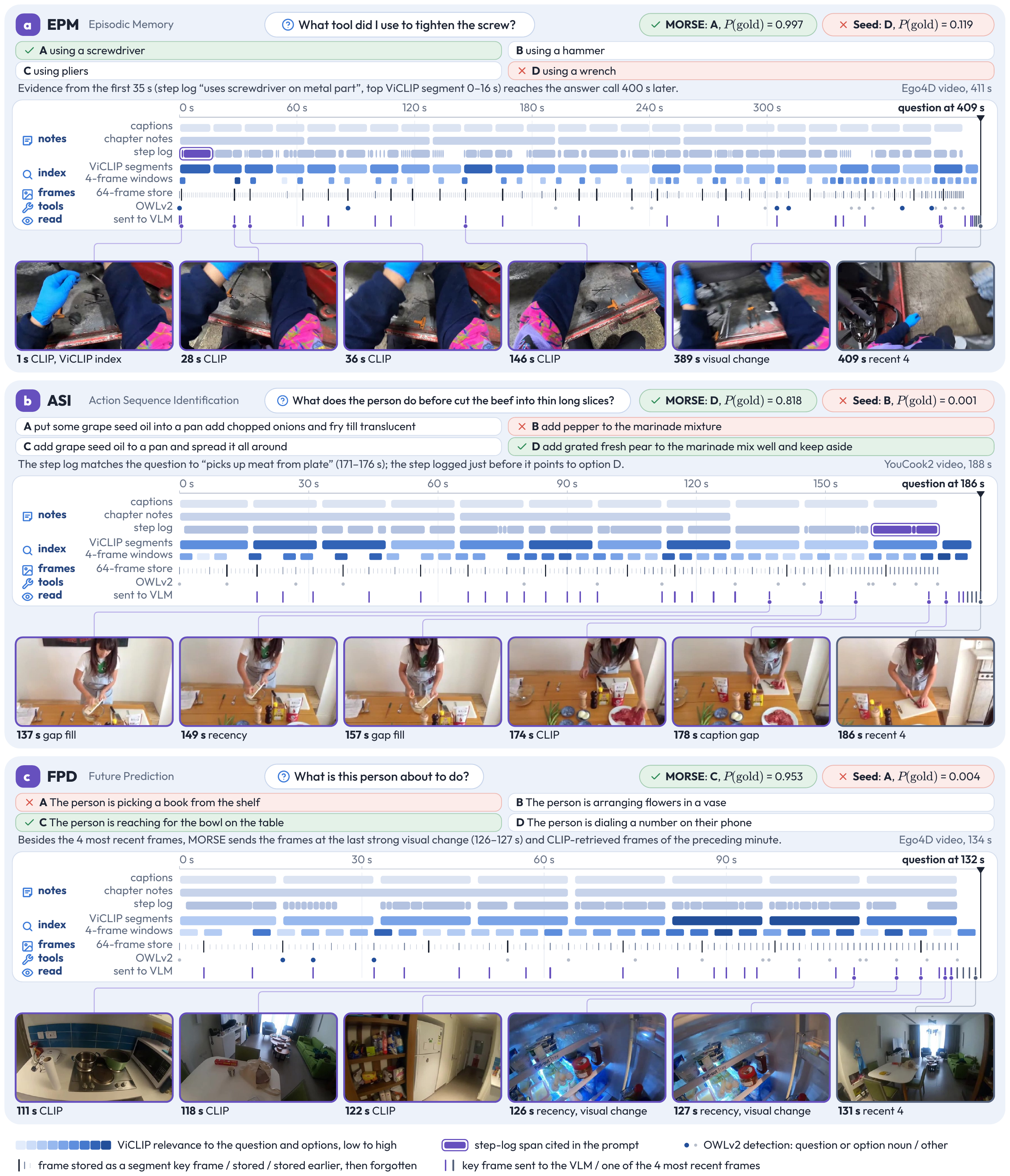}
\caption{\textbf{Memory of the final harness at question time on three OVO-Bench test questions} (tasks EPM, ASI,
and FPD). Questions and options are reproduced verbatim. Each panel gives the question, its options (green: correct;
red: the seed's choice), and both answers with
the probability $P(\text{gold})$ of the correct option. The timeline shows what the write path stored from the start
of the video to the question (black marker). \emph{Notes}: captions every 16\,s, chapter notes over four captions,
and a step log of who does what (purple outline: steps cited in the answer prompt). \emph{Index}: ViCLIP vectors of
16-s segments and 4-frame windows, darker when closer to the question and options. \emph{Frames}: the 64-frame store
(tall ticks: segment key frames; pale ticks: frames since forgotten). \emph{Tools}: OWLv2 detections, dark if the
noun is named in the question or options. \emph{Read}: the frames sent to the VLM, retrieved key frames in purple and
the 4 most recent frames in gray. Thumbnails show six of them with the selector that chose each: similarity to the
question (\emph{CLIP}, \emph{ViCLIP index}), temporal coverage (\emph{recency}, \emph{gap fill}, \emph{caption gap}),
or the last strong scene change (\emph{visual change}). The seed sends only the captions and the 4 most recent
frames.}
\label{fig:memory_examples}
\end{figure*}

\Cref{fig:memory_examples} shows what the final harness $H_{\mathrm{S11}}$ holds when a question arrives and what it
sends to the answering VLM, for three OVO-Bench test questions that $H_{\mathrm{S11}}$ answers correctly and the seed
harness does not. The questions cover a backward-tracing episodic question asked 400\,s after the evidence, a step-order
question, and a real-time prediction. Every row of a timeline is built on the write path while the video streams:
host captions every 16\,s, chapter notes over four captions, the step log, the ViCLIP segment and window index, the
64-frame store with its forgetting, and OWLv2 detections. At question time the harness only reads these structures,
ranks the index against the question and options, and makes one VLM call.

\vspace{3pt} \noindent \textbf{(a) Episodic memory (EPM).} The screwdriver is used only in the first 35\,s of a 411-s video,
and the question arrives at 409\,s. Of the segment vectors written while the video streamed, the ViCLIP index ranks
0--16\,s highest for the options; CLIP frame retrieval adds frames at 1, 28, and 36\,s; and the write-time step log
holds ``uses screwdriver on metal part'' (2--16\,s). The answer call therefore sees the evidence 400\,s after it streamed
past ($P(\text{gold})=0.997$), whereas the seed, which sees only the captions and the last four frames, answers
``wrench''.

\vspace{3pt} \noindent \textbf{(b) Action sequence identification (ASI).} The write-time step log records who does what with one VLM call per
16\,s and merges repeated actions into spans. At answer time the harness matches the question's step to ``picks up meat
from plate'' (171--176\,s) by CLIP text similarity, compares the steps logged just before it (``mixes contents in
bowl'', ``puts spoon down'', 161--171\,s) with the options, and adds a note naming option D to the prompt
($P(\text{gold})=0.818$). The seed answers B.

\vspace{3pt} \noindent \textbf{(c) Future prediction (FPD).} Both harnesses see the four most recent frames, in which the
camera wearer walks from the open fridge past a shelf toward the dining table. \method{} also sends the frames at the
last strong change of its per-second CLIP trace (126--127\,s) with a note giving that time, and CLIP-retrieved frames
from the preceding minute in the kitchen; it predicts reaching for the bowl ($P(\text{gold})=0.953$). The seed, without
key frames, predicts taking a book from the shelf.

\end{document}